\documentclass{article}

\usepackage[final,eandd]{neurips_2026}
\usepackage[utf8]{inputenc}
\usepackage[T1]{fontenc}
\usepackage{hyperref}
\usepackage{url}
\usepackage{xurl}
\usepackage{booktabs}
\usepackage{amsfonts}
\usepackage{nicefrac}
\usepackage{microtype}
\usepackage{xcolor}
\usepackage{graphicx}
\usepackage{caption}
\usepackage{placeins}
\usepackage{amsmath,amssymb}
\usepackage{mathtools}
\usepackage{amsthm}
\usepackage{bbm}
\usepackage{bm}
\usepackage{subfigure}
\usepackage{float}
\usepackage{adjustbox}
\usepackage{colortbl}
\usepackage{arydshln}
\usepackage{algorithm,algpseudocode}
\usepackage{algorithmicx}
\usepackage{wrapfig}
\usepackage{pifont}
\usepackage{bbding}
\usepackage{xspace}
\usepackage{multicol}
\usepackage{array}
\usepackage{multirow}
\usepackage{pgf}
\usepackage{enumitem}
\usepackage{hyperref}
\usepackage{url}
\usepackage{booktabs}
\usepackage{amsmath,amssymb}
\usepackage{graphicx}
\usepackage{xcolor}
\usepackage{array}
\usepackage{tabularx}
\usepackage{longtable}
\usepackage{adjustbox}
\usepackage{enumitem}
\usepackage{xspace}
\usepackage{caption}

\newcolumntype{Y}{>{\raggedright\arraybackslash}X}
\newcolumntype{P}[1]{>{\raggedright\arraybackslash}p{#1}}
\setlist[itemize]{leftmargin=1.2em,itemsep=1pt,topsep=2pt}
\setlist[enumerate]{leftmargin=1.4em,itemsep=1pt,topsep=2pt}
\definecolor{myblue}{HTML}{76C7E5}
\newcommand{\heatcell}[3]{%
    \pgfmathsetmacro{\percent}{(#1 - #2) / (#3 - #2) * 50 + 10}%
    \edef\temp{\noexpand\cellcolor{myblue!\percent}\relax \noexpand\makebox[2.2em][r]{#1}}%
    \temp
}
\newcommand{\heatcellb}[3]{%
    \pgfmathsetmacro{\percent}{(#1 - #2) / (#3 - #2) * 50 + 10}%
    \edef\temp{\noexpand\cellcolor{myblue!\percent}\relax \noexpand\makebox[2.2em][r]{\noexpand\textbf{#1}}}%
    \temp
}
\newcommand{\cmark}{\textcolor{green!60!black}{\ding{51}}}
\newcommand{\xmark}{\textcolor{red}{\ding{55}}}
\newcommand{\ours}{\textsc{MedHorizon}\xspace}
\newcommand{\figref}[1]{Fig.~\ref{#1}}
\newcommand{\tabref}[1]{Tab.~\ref{#1}}

\makeatletter
\DeclareRobustCommand\onedot{\futurelet\@let@token\@onedot}
\def\@onedot{\ifx\@let@token.\else.\null\fi\xspace}

\makeatother

\title{\ours: Towards Long-context Medical Video Understanding in the Wild}

\author{%
  Bodong Du\textsuperscript{1,\ensuremath{\dagger}} \quad
  Bowen Liu\textsuperscript{1,\ensuremath{\dagger},*} \quad
  Yang Yu\textsuperscript{1} \quad
  Xinpeng Ding\textsuperscript{3} \\
  \bfseries Zhiheng Wu\textsuperscript{2} \quad
  Shuning Wang\textsuperscript{2} \quad
  Shuo Nie\textsuperscript{2} \quad
  Naiming Liu\textsuperscript{2} \\
  \bfseries Qifeng Chen\textsuperscript{1} \quad
  Yangqiu Song\textsuperscript{1} \quad
  Xiaomeng Li\textsuperscript{1,\ding{41}} \\[4pt]
  \normalfont\textsuperscript{1}The Hong Kong University of Science and Technology \\
  \normalfont\textsuperscript{2}Baidu Inc. \quad
  \textsuperscript{3}Xidian University
}

\begin{document}

\maketitle
\begingroup
  \renewcommand{\thefootnote}{}
  \footnotetext{%
    \textsuperscript{\ensuremath{\dagger}}Equal Contribution \quad
    \textsuperscript{*}Project Leader \quad
    \textsuperscript{\ding{41}}Corresponding Author.}
\endgroup

\begin{abstract}
Medical multimodal large language models (MLLMs) have advanced image understanding and short-video analysis, but real clinical review often requires full-procedure video understanding. Unlike general long videos, medical procedures contain highly redundant anatomical views, while decisive evidence is temporally sparse, spatially subtle, and context dependent. Existing benchmarks often assume this evidence has already been localized through images, short clips, or pre-segmented videos, leaving the retrieval-before-reasoning problem under-tested. We introduce \ours, an in-the-wild benchmark for long-context medical video understanding. \ours preserves 759 hours of full-length clinical procedures and provides 1,253 evidence-grounded multiple-choice questions that jointly evaluate sparse evidence understanding and multi-hop clinical reasoning. Its evidence is extremely sparse, with only 0.166\% evidence frames on average, requiring models to search noisy procedural streams before interpreting and aggregating findings. We evaluate representative general-domain, medical-domain, and long-video MLLMs. The best model reaches only 41.1\% accuracy, showing that current systems remain far from robust full-procedure understanding. Further analysis yields four key findings: performance does not scale reliably with more frames, evidence retrieval and clinical interpretation remain primary bottlenecks; these bottlenecks are rooted in weak procedural reasoning and attention drift under redundancy, and generic sampling methods only partially balances local detail with global coverage. \ours provides a rigorous testbed for MLLMs that retrieve sparse evidence and reason over complete clinical workflows.
\end{abstract}

\section{Introduction}


Long medical video understanding plays a vital role in diagnostic assessment, procedural quality control, surgical workflow understanding, and clinical decision-making. In endoscopic examination, clinicians assess mucosal inspection and bowel preparation across the withdrawal phase ~\cite{asge2024quality, corley2014adr}; in obstetric ultrasound, standard views and anatomical coverage must be verified throughout the examination~\cite{salomon2022isoug}; and in surgery, phases and critical events may be separated by long temporal gaps~\cite{demir2023workflow}. These scenarios require \textit{\textbf{full-length clinical video understanding}} rather than isolated images or curated short clips. Recent medical multimodal large language models (MLLMs) have advanced image understanding and short-video analysis, making it timely to evaluate whether they can meet this full-procedure requirement.

\begin{figure}[t]
    \centering
    \includegraphics[width=\linewidth]{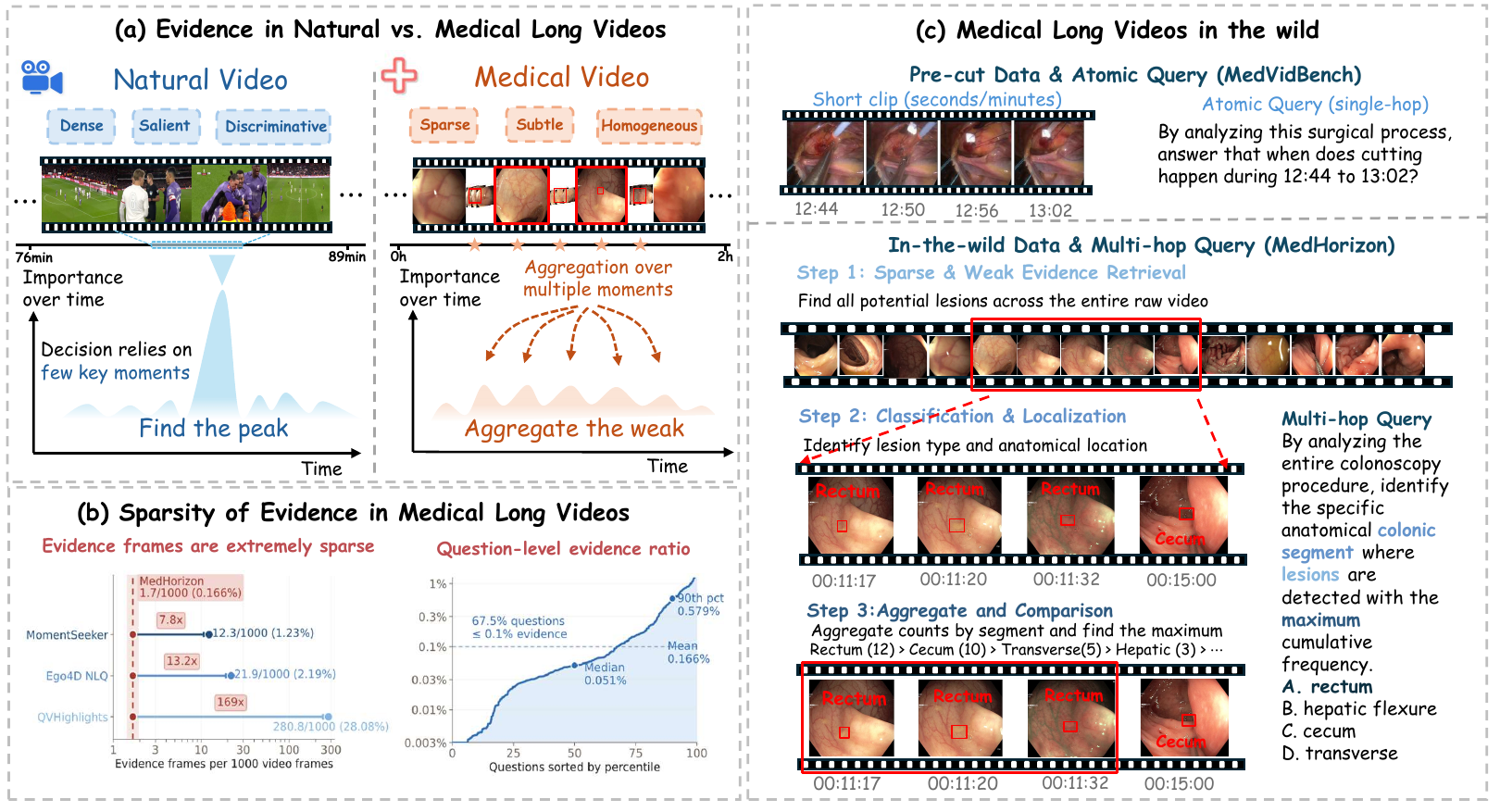}
    \caption{\ours targets the distinctive evidence structure of long medical videos. (a) Unlike many natural videos, where a few discriminative moments can answer a query, medical procedure videos contain highly redundant frames and weak evidence distributed over time. (b) This evidence is extremely sparse: \ours contains 1.7 evidence frames per 1,000 video frames on average, with a mean question-level evidence ratio of 0.166\%. (c) \ours evaluates full-procedure retrieval and multi-hop clinical reasoning rather than pre-cut clip QA~\cite{medgrpo}.}
    \label{fig:main_teaser}
\end{figure}

These full-procedure videos define a \textit{\textbf{distinctive long-context medical setting}}. 
Unlike general long videos~\cite{videomme2024,longvideobench2024,wang2025lvbench}, where query-relevant evidence is often relatively \emph{dense}, \emph{salient}, and visually \emph{discriminative}, medical long videos are characterized by \textbf{\emph{sparse}} evidence, \textbf{\emph{subtle}} spatial findings, and highly \textbf{\emph{homogeneous}} anatomical appearances across adjacent frames, as shown in Fig.~\ref{fig:main_teaser}(a). 
In such videos, decisive cues may appear only within short temporal windows and can be obscured by strong visual redundancy, while their interpretation often depends on anatomical location, procedural phase, or cross-moment observations. 
Thus, the challenge is not merely length, but the combination of temporally sparse evidence, spatially subtle findings, high frame similarity, and strong context dependence.

Existing medical video benchmarks~\cite{he2024pitvqa, li2024ophnet, thapa2025openbiomedvid, qian2026sureon} capture this challenge only partially. Many evaluations assume query-relevant evidence has already been localized, because inputs are images, short clips, pre-segmented videos, or questions answerable from limited visual context. Such settings under-test a core capability in real clinical review: \textbf{finding weak evidence before reasoning over it}. Models must therefore search long, visually similar procedural streams, identify rare clinically relevant moments, bind them to anatomical or temporal context, and aggregate observations before answering.

To address this gap, we introduce \textbf{\ours}, a benchmark for \textbf{in-the-wild long-context medical video understanding}. \ours comprises 340 public full-length videos from 8 datasets, spanning 7 organs and 2 clinical scenarios: diagnostic examination and surgery. It contains \textbf{759 hours} of clinical video, with procedures lasting up to 37.2 hours. Rather than trimming videos into answer-containing clips, \ours preserves the full procedural stream and evaluates models under complete video context. This makes it a testbed where models operate over noisy, redundant, and weakly discriminative visual records rather than curated snippets. Built on these full-context videos, \ours provides 1,253 rigorously verified multiple-choice questions that jointly test sparse evidence perception and multi-hop semantic reasoning. Each question is anchored to full-video evidence and refined through distractor strengthening, language rewriting, and human verification, reducing shortcut solutions based on superficial textual cues or isolated visual priors. Models must therefore locate sparse evidence, interpret subtle local details, connect them with clinical context, and reason across observations before selecting an answer.

\ours stresses in-the-wild medical video-query understanding along both video and query axes. On the video axis, decisive evidence is extremely sparse in redundant streams: \ours has only \textbf{0.166\%} evidence frames, or 1.7 per 1,000 frames, \textbf{$\textbf{169}\times$ sparser} than general benchmarks such as QVHighlights~\cite{lei2021detecting}, requiring extreme needle-in-a-haystack localization precision, as shown in Fig.~\ref{fig:main_teaser}(b). On the query axis, many questions go beyond single-frame or short-clip perception. For example, a lesion-distribution query may require retrieving candidate lesions across the full procedure, localizing them to anatomical regions, filtering visually similar distractors, and aggregating findings before the final answer, as shown in Fig.~\ref{fig:main_teaser}(c).

We evaluate a broad set of models on \ours, covering general-domain MLLMs, medical-domain MLLMs, and specialized long-video methods. The best model reaches only \textbf{41.1\% accuracy}, showing that current models remain far from robust in-the-wild long-context medical video understanding. Further analysis yields \textbf{four key findings}: non-monotonic frame scaling, the bottlenecks in clinical evidence retrieval and interpretation, the underlying mechanisms of weak procedural reasoning and attention drift that explain these bottlenecks, and the limited ability of generic long-video sampling methods. These results indicate that simply scaling context length is insufficient; future medical MLLMs need stronger mechanisms for sparse evidence perception and contextual reasoning. In summary, our contributions are threefold:
\begin{itemize}
[leftmargin=*, labelsep=0.5em, itemsep=0.2em]
    \item[\ding{182}] \textbf{Problem framing.} We identify full-procedure medical video understanding as a distinct long-video setting where evidence is sparse, weak, visually redundant, and distributed over time, unlike the highly discriminative moments often used in general long-video settings.
    \item[\ding{183}] \textbf{Benchmark artifact.} We introduce \ours, a benchmark of 759 hours of raw clinical procedures and 1,253 evidence-grounded questions that preserve full procedural context and require sparse retrieval, clinical binding, and multi-hop aggregation.
    \item[\ding{184}] \textbf{Diagnostic findings.} We evaluate representative general and medical MLLMs, revealing four key findings: non-monotonic frame scaling, persistent bottlenecks in retrieval and interpretation, the weak procedural reasoning and attention drift, and the limited ability of generic sampling strategies.
\end{itemize}

\section{Related Work}

\subsection{General Long Video Understanding}

General long-video understanding has advanced rapidly in non-clinical settings. Recent video MLLMs, including InternVL, Qwen3-VL, MiniCPM-V, and LLaVA-Video, extend the processable horizon through sparse temporal input, hierarchical aggregation, and efficient attention~\cite{internvl,internvl3,internvl3.5,qwen2vl,qwen2.5vl,qwen3vl,minicpmv26,zhang2024llavanextvideo}. Open-domain benchmarks such as Video-MME, MVBench, and LongVideoBench, together with efficient samplers such as AKS, ViLAMP, and wavelet-based frame selection, now provide a strong testbed for long-form video reasoning under constrained compute budgets~\cite{videomme2024,mvbench2024,longvideobench2024,AKS,vilamp,chen2026wavelet}. However, these methods and benchmarks target natural-video semantics rather than clinical workflows. They usually assume that informative evidence is visually salient or recoverable by representative sampling, whereas long medical videos often contain sparse but decisive findings scattered across lengthy procedures. Consequently, success on general long-video benchmarks does not directly translate to long-context medical video understanding.

\subsection{Medical Video Benchmarks}

Existing medical video benchmarks are closer to our target setting, but the dominant evaluation mode remains clip-level or short-horizon. PitVQA, OphNet, OpenBiomedVid, and SUREON study surgical or biomedical video QA and reasoning~\cite{he2024pitvqa,li2024ophnet,thapa2025openbiomedvid,qian2026sureon}; SurgPub-Video improves surgical VQA scale, and MedGRPO introduces MedVidBench with heterogeneous medical videos and multi-granular supervision~\cite{surgpubvideo,medgrpo}. These resources are important, but they mostly focus on bounded clips or videos up to 30 minutes rather than ultra-long full-procedure evaluation. They therefore broaden medical video supervision without fully testing retrieval and reasoning over sparse evidence in complete examinations. This gap matters because real clinical interpretation is inherently process-level. Colonoscopy quality depends on complete inspection and withdrawal behavior over the whole procedure~\cite{asge2024quality,corley2014adr,duloy2019video,decarvalho2023withdrawal,liu2026divide}; ultrasound assessment depends on sufficient standard views and cine loops across the study~\cite{salomon2022isoug,lang2015ase,dormagen2015cine}; and surgical understanding requires tracking phases, steps, and events over time~\cite{demir2023workflow,carstens2025surgicalai}. \ours complements existing benchmarks by explicitly evaluating this full-procedure setting, where a model must both localize sparse clinical evidence and integrate temporally separated observations into a coherent judgment.

\begin{figure}
    \centering
    \includegraphics[width=0.9\linewidth]{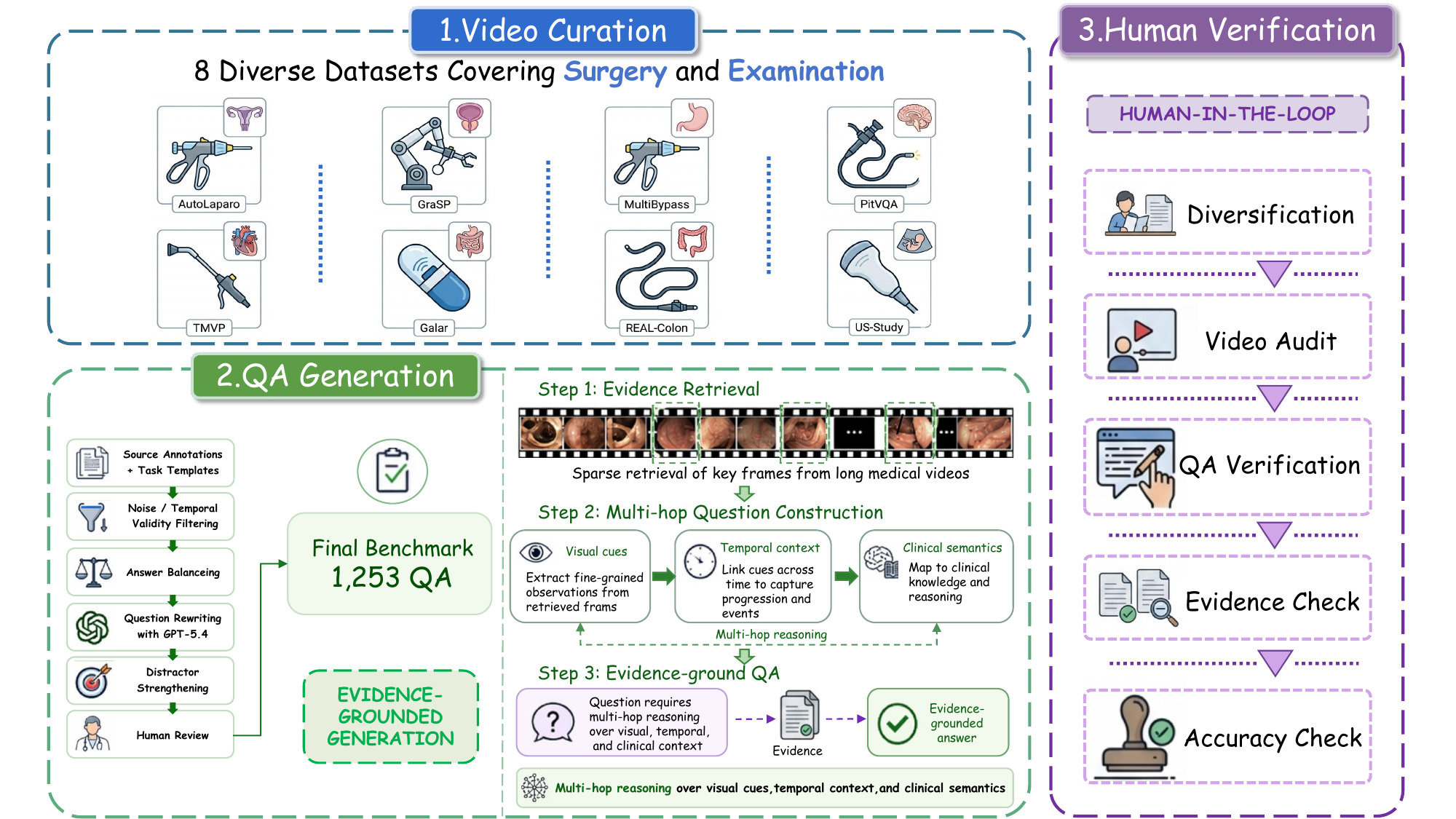}
    \caption{Benchmark construction pipeline.}
   \label{fig:pipeline}
\end{figure}

\begin{figure*}[ht]
    \centering
    \subfigure[Video Category.]{
    \centering
        \includegraphics[width=0.25\linewidth]{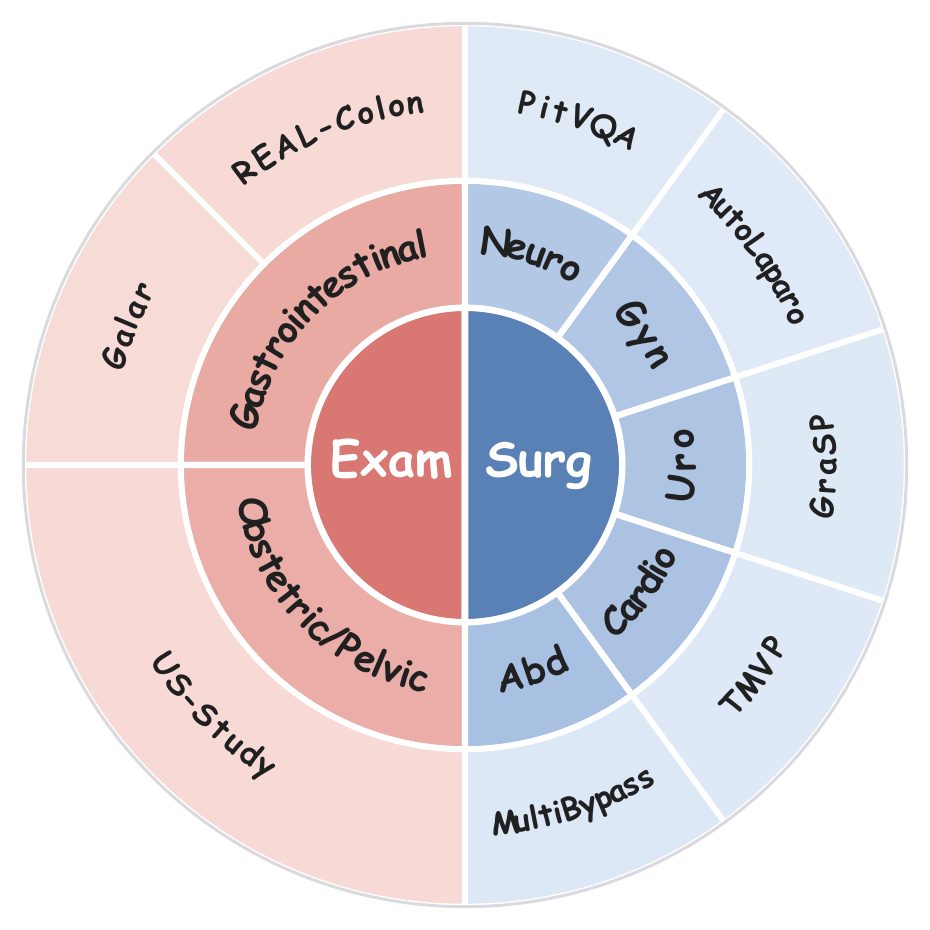}
        \label{fig:data_a}
    }
    \hfill
        \subfigure[Video Duration Distribution]{
        \includegraphics[width=0.4\linewidth]{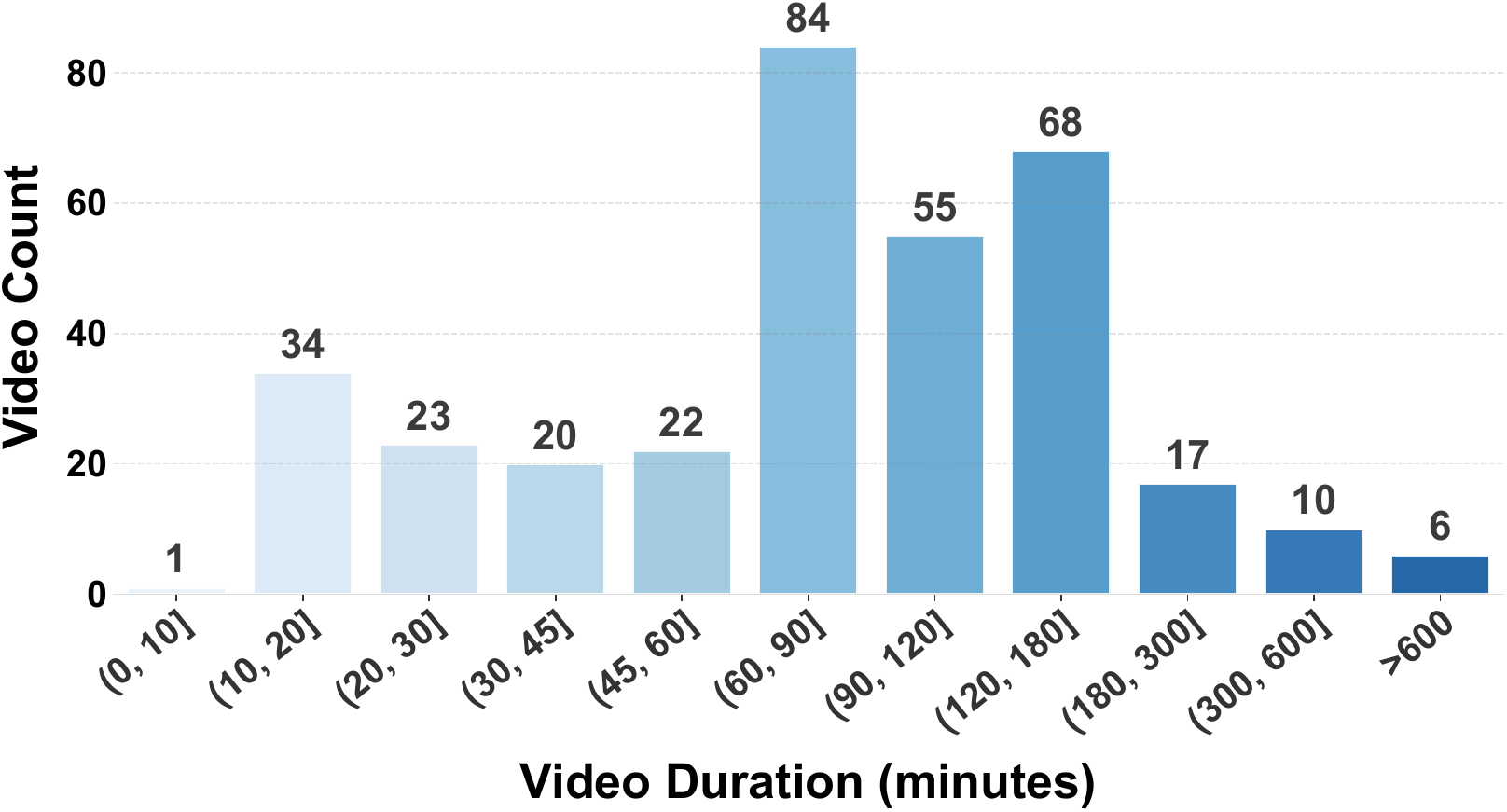}
        \label{fig:data_b}
    }
    \hfill
    \subfigure[QA density]{
        \includegraphics[width=0.25\linewidth]{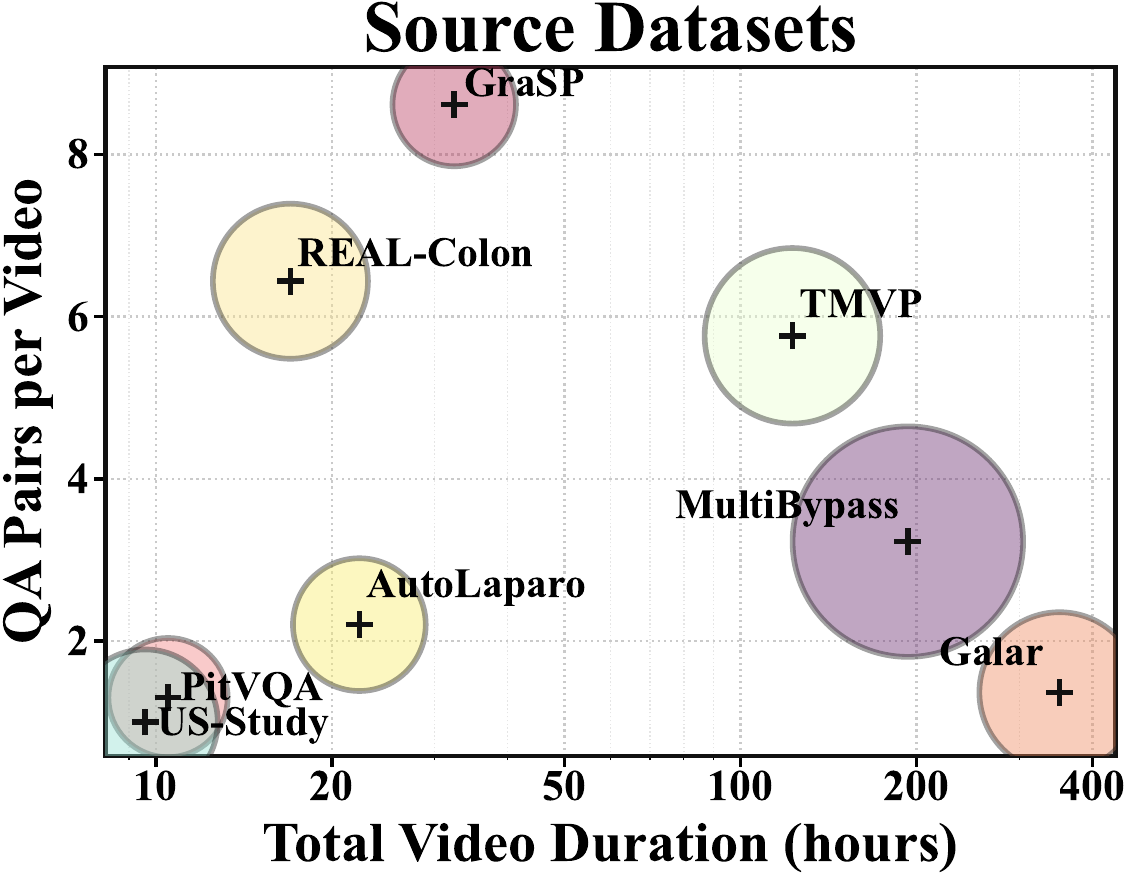}
        \label{fig:data_c}
    }
    \hfill
    \caption{Overview of the source corpus.}
    \label{fig:data_overview}
\end{figure*}



\setlength{\tabcolsep}{3pt}

\section{Benchmark Construction}
\label{sec:dataset}

\subsection{Data Curation}

\ours is built from 8 public medical video sources, totaling 340 videos and 759 decoded hours across 7 organs and 2 clinical scene types, including surgery, colonoscopy, capsule endoscopy, and ultrasound, as summarized in \figref{fig:data_overview}(a). This breadth prevents long-context medical video understanding from collapsing to a single procedure family or annotation style, and forces transfer across distinct visual appearances, workflows, and temporal rhythms, including cardiac and abdominal surgery, gastrointestinal examination, and obstetric ultrasound.

The source corpus is deliberately long-tailed in duration: \figref{fig:data_overview}(b) shows many videos last over one hour, with the longest study reaching 37.2 hours. This matters because realistic deployment retrieval bottlenecks come from searching full procedural streams, not pre-trimmed clips. As illustrated in \figref{fig:data_overview}(c), question density varies considerably across datasets, primarily driven by discrepancies in temporal granularity and clinical scope among released annotations. Consequently, we decouple corpus coverage from evaluation depth in our subsequent construction phase.

\subsection{Construction Pipeline}

\begin{figure}[t]
    \centering
    \includegraphics[width=\linewidth]{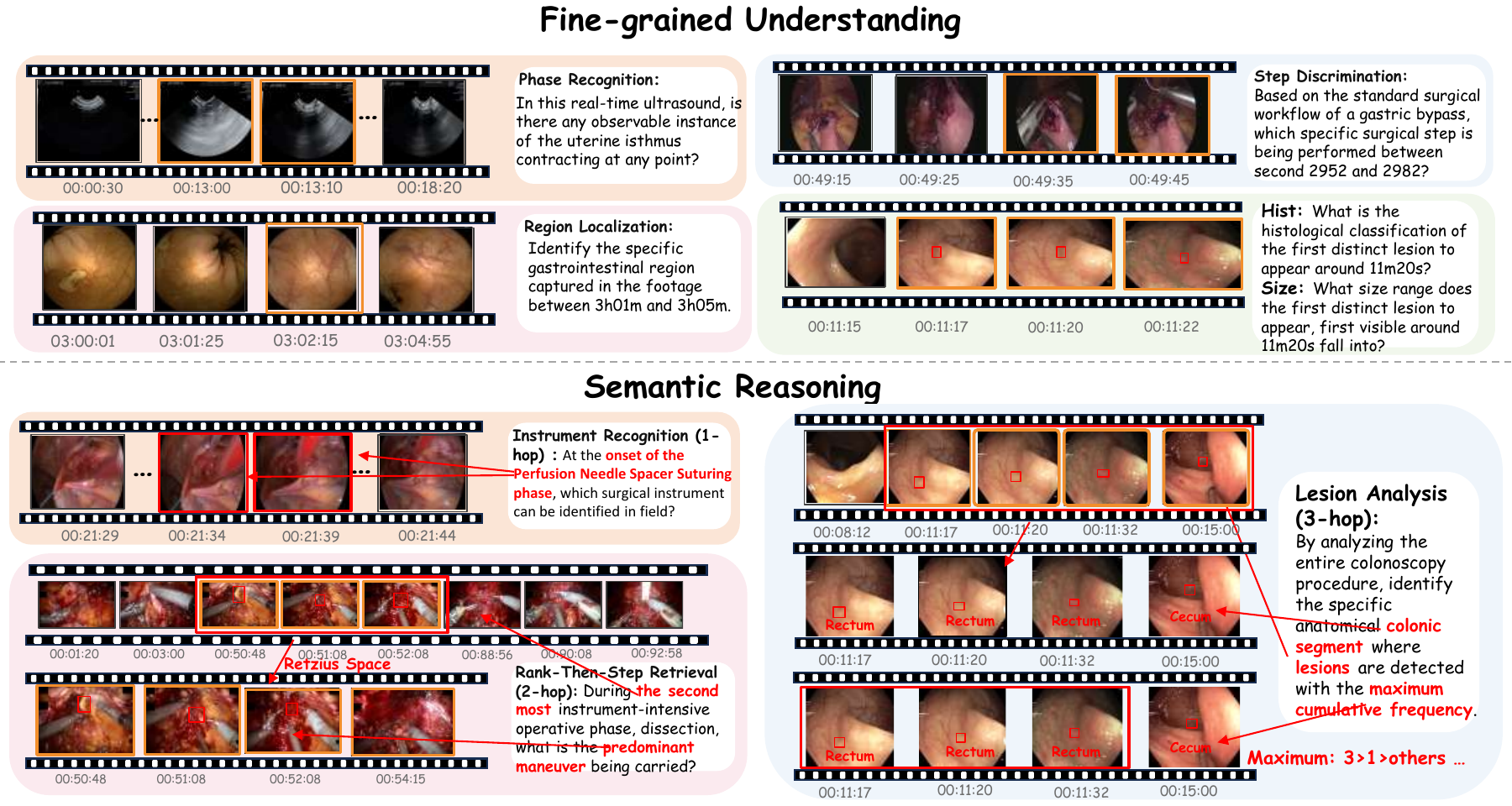}
    \caption{Representative video-question pairs across fine-grained understanding and semantic reasoning tasks in \ours.}
    \label{fig:all_demo}
\end{figure}

The benchmark is constructed through the evidence grounded pipeline in \figref{fig:pipeline}. Starting from diverse surgical and examination videos, we collect source annotations and task schemas covering phases,  anatomical regions, examination targets, lesion properties, instruments, and procedure level events. 

In the second step, each QA item is generated from explicit video evidence in expert annotations. The process retrieves evidence from the full video, including sparse key frames and temporally relevant segments, then builds candidate questions around the annotated event or finding. We filter noisy, ambiguous, or temporally inconsistent candidates, balance answer distributions, rewrite questions into direct natural language with GPT 5.4, and strengthen distractors using nearby temporal events and clinically plausible alternatives. This design gives every retained question a traceable visual basis and challenging distractors without unsupported claims. 

The final construction stage, human in the loop verification, uses human review as an audit layer. In each construction iteration, all items went through four checks: whether the video is usable for the intended task, whether the question and answer choices are well formed, whether the selected evidence truly supports the answer, and whether the final label is consistent with the original annotation. Items failing these checks are revised or removed, keeping the benchmark grounded in observable video content rather than post hoc textual assumptions. Detailed audit statistics, construction funnel counts, evidence export fields, and the oracle evidence protocol are provided in the appendix \ref{sec:supp_benchmark_card}.


\subsection{Dataset Statistics}

\begin{wrapfigure}{r}{0.45\linewidth}
\vspace{-2em}
  \centering
  \includegraphics[width=\linewidth]{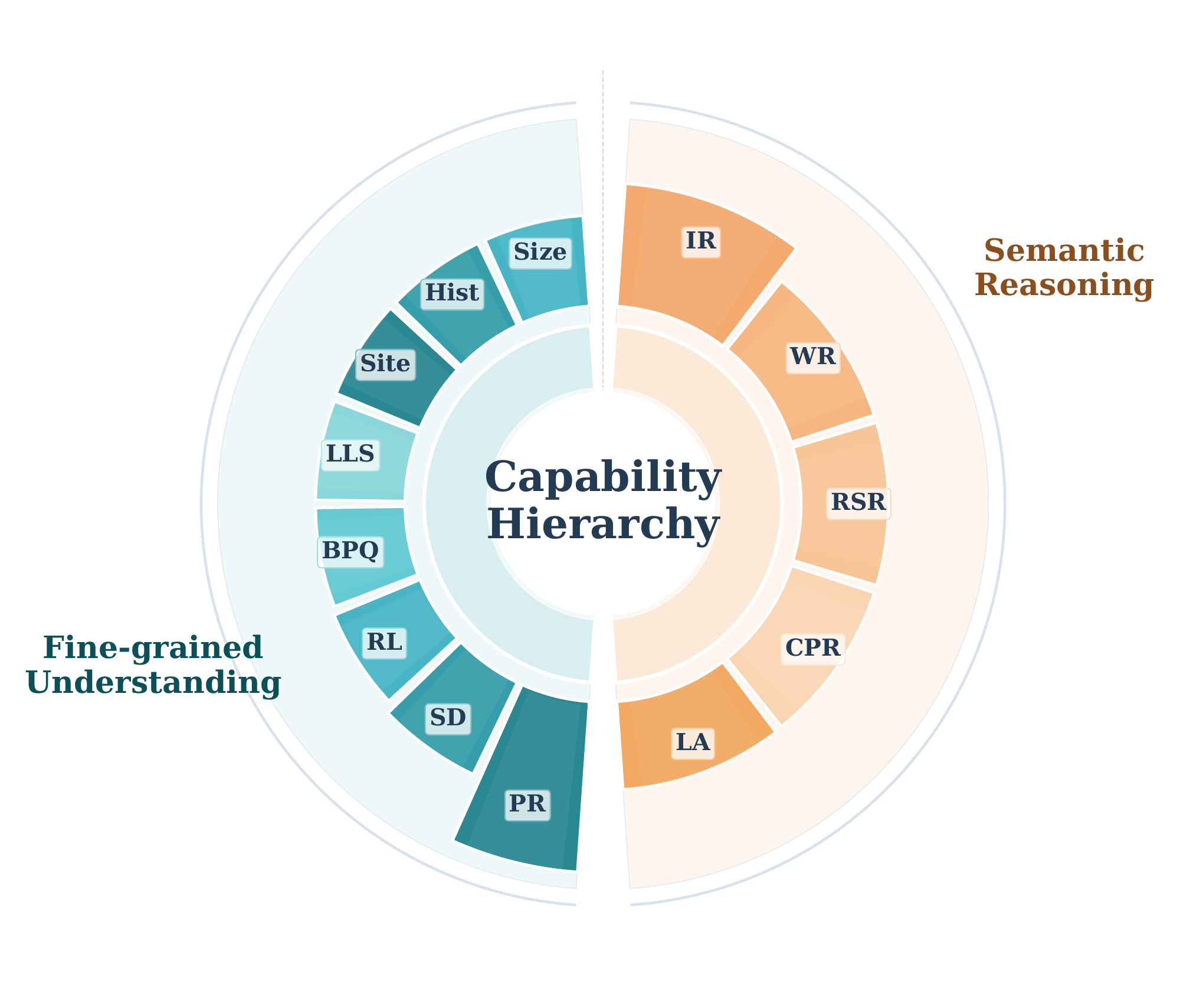}
  \caption{Capability taxonomy.}
  \label{fig:capability_taxonomy}

\end{wrapfigure}
\ours is distinguished by its emphasis on evidence-based reasoning over raw, long-form clinical videos. As illustrated in  \figref{fig:all_demo} and \figref{fig:capability_taxonomy}, we categorize the benchmark into two primary dimensions: fine-grained understanding and multi-hop semantic reasoning. This hierarchy requires models to move beyond single-label recognition toward searching and integrating clinically meaningful evidence that is often sparse, temporally distant, and visually subtle. 
\textbf{Fine-grained understanding} tasks focus on the localization and identification of localized visual cues. This includes procedural phase recognition (PR), step discrimination (SD), anatomical region localization (RL), and assessment of bowel preparation quality (BPQ). Furthermore, "binding" tasks (Site, Hist, Size) and largest-lesion identification (LLS) require models to associate specific lesions with their respective anatomical, histological, or morphological attributes. \textbf{Semantic reasoning} tasks are further stratified by reasoning depth to measure cross-temporal integration. At the foundational level, \textbf{one-hop visual-semantic} tasks, such as Instrument Recognition (IR), require the model to map localized appearances to functional categories. Moving toward higher complexity, \textbf{two-hop} comparative tasks, including Workload Ranking (WR) and Ranked Step Retrieval (RSR), demand that models first identify procedural cues or steps across the timeline and subsequently perform relative comparison or temporal ranking. The highest level of complexity is captured by \textbf{three-hop} aggregative tasks, such as Count-Proportion Reasoning (CPR) and Lesion Analysis (LA). These require a sophisticated pipeline of locating repeated events or lesion-relevant moments, binding them to multi-dimensional contexts spanning appearance, location, and time, and finally aggregating these disparate observations into a unified, procedure-level clinical judgment.


Finally, \tabref{tab:recent_compare} situates \ours relative to recent medical video benchmarks. Existing resources often focus on either surgery or examination, rely on short clips or pre-selected segments, or flatten evaluation into atomic recognition. By contrast, \ours is the only benchmark in this comparison that jointly preserves full-procedure integrity, evaluates sparse evidence retrieval, and requires multi-hop reasoning across multiple clinical scenes.

\begin{table}[t]
\centering
\caption{Comparison with recent medical video benchmarks. }
\label{tab:recent_compare}
\scriptsize
\setlength{\tabcolsep}{3pt}
\resizebox{0.9\linewidth}{!}{%
\begin{tabular}{l c c c c c c c}
\toprule
Dataset & Year & Surg. & Exam. & Long-context & Extremely sparse & Multi-hop & Total(h) \\
\midrule
PitVQA~\cite{he2024pitvqa} & 2024 & \cmark & \xmark & \xmark & \xmark & \xmark & 2.1 \\
OphNet~\cite{li2024ophnet} & 2024 & \cmark & \xmark & \cmark & \xmark & \xmark & 24.5 \\
SurgeryVideoQA~\cite{thapa2025openbiomedvid} & 2025 & \cmark & \xmark & \xmark & \xmark & \xmark & 1.8 \\
MIMICEchoQA~\cite{mimicechoqa2025} & 2025 & \xmark & \cmark & \xmark & \xmark & \xmark & 5.2 \\
SurgPub-Video~\cite{surgpubvideo} & 2026 & \cmark & \xmark & \cmark & \xmark & \xmark & 31.5 \\
MedVidBench~\cite{medgrpo} & 2026 & \cmark & \cmark & \xmark & \xmark & \cmark & 22.4 \\
SUREON~\cite{qian2026sureon} & 2026 & \cmark & \xmark & \xmark & \xmark & \xmark & 3.0 \\
\midrule
\textbf{\ours (Ours)} & 2026 & \cmark & \cmark & \cmark & \cmark & \cmark & \textbf{759.0} \\
\bottomrule
\end{tabular}
}
\end{table}

\section{Experiments}
\label{sec:experiments}

We evaluate a diverse set of state-of-the-art MLLMs, grouped into closed-source MLLMs, general open-source video MLLMs, specialized long-video methods, and medical MLLMs. For closed-source models with native video input, we sample videos at 1 FPS and cap the input at 1,248 frames to ensure stable API inference on very long videos. For general open-source video MLLMs, models with native long-video support use 1 FPS decoding with the maximum-frame cap from their official implementations, while others use 64 uniformly sampled frames. For specialized long-video methods, we follow their official frame budgets and sampling strategies. For medical MLLMs, all evaluated models support native long-video inference, so we use 1 FPS decoding with their official maximum-frame caps. All models are evaluated with official settings and run on 8 A800.

Four research questions organize the analysis. \textbf{RQ1:} Does long-context medical video understanding follow a frame-scaling law as general domains? \textbf{RQ2:} What limits the performance of MLLMs in long-context medical video understanding? \textbf{RQ3:} What underlying mechanisms explain such bottlenecks? \textbf{RQ4:} Can generic sampling strategies address the unique challenges of long-context medical  videos understanding? 







\subsection{Overall Performance}

\begin{table*}[ht]
\renewcommand{\arraystretch}{1.18}
\centering
\caption{Main results on \ours. Input reports the visual input protocol: a number indicates a fixed uniform sampled frame budget, while ``1 FPS'' indicates fps-based video sampling. Best results are in \textbf{bold}. Darker shades indicate stronger values within each column.}
\label{tab:main_results}
\begin{adjustbox}{width=1\linewidth}
\scriptsize
\setlength{\tabcolsep}{2.7pt}
\begin{tabular}{l c c c c c c c c c c c c c c c c}
\toprule
\multirow{2}{*}{\textbf{Models}} & \multirow{2}{*}{\textbf{Size}} & \multirow{2}{*}{\textbf{Input}} & \multirow{2}{*}{\textbf{Ov.}} & \multicolumn{8}{c}{\textbf{Fine-grained Understanding}} & \multicolumn{5}{c}{\textbf{Semantic Reasoning}} \\
\cmidrule(lr){5-12}\cmidrule(lr){13-17}
 &  &  &  & PR & SD & RL & BPQ & LLS & Site & Hist & Size & IR & WR & RSR & CPR & LA \\

\midrule
\midrule
\rowcolor[HTML]{E3FDFD}
\multicolumn{17}{c}{\bfseries\itshape Closed-Source MLLMs} \\
\midrule
GPT-5.4-mini~\protect\cite{gpt54mini} & -- & 1 FPS & \heatcell{31.2}{18.2}{41.1} & \heatcell{35.0}{17.0}{38.9} & \heatcell{17.0}{8.0}{40.2} & \heatcell{58.3}{11.7}{70.0} & \heatcell{38.5}{10.3}{59.0} & \heatcell{0.0}{0.0}{61.5} & \heatcell{17.0}{8.5}{31.9} & \heatcell{19.1}{0.0}{55.3} & \heatcell{17.0}{0.0}{48.9} & \heatcell{34.5}{12.3}{62.6} & \heatcell{42.3}{11.5}{61.5} & \heatcell{57.7}{19.2}{69.2} & \heatcell{11.5}{3.8}{42.3} & \heatcell{18.8}{0.0}{37.5} \\
GPT-5.4~\protect\cite{gpt54} & -- & 1 FPS & \heatcell{32.6}{18.2}{41.1} & \heatcell{36.4}{17.0}{38.9} & \heatcell{20.5}{8.0}{40.2} & \heatcell{61.7}{11.7}{70.0} & \heatcell{41.0}{10.3}{59.0} & \heatcell{20.5}{0.0}{61.5} & \heatcell{29.8}{8.5}{31.9} & \heatcell{21.3}{0.0}{55.3} & \heatcell{31.9}{0.0}{48.9} & \heatcell{23.4}{12.3}{62.6} & \heatcell{38.5}{11.5}{61.5} & \heatcellb{69.2}{19.2}{69.2} & \heatcell{30.8}{3.8}{42.3} & \heatcell{18.8}{0.0}{37.5} \\
Gemini-3.1-Pro-Preview~\protect\cite{gemini3propreview} & -- & 1 FPS & \heatcellb{41.1}{18.2}{41.1} & \heatcell{33.3}{17.0}{38.9} & \heatcell{31.2}{8.0}{40.2} & \heatcell{40.0}{11.7}{70.0} & \heatcellb{59.0}{10.3}{59.0} & \heatcellb{61.5}{0.0}{61.5} & \heatcell{21.3}{8.5}{31.9} & \heatcell{31.9}{0.0}{55.3} & \heatcellb{48.9}{0.0}{48.9} & \heatcellb{62.6}{12.3}{62.6} & \heatcellb{61.5}{11.5}{61.5} & \heatcellb{69.2}{19.2}{69.2} & \heatcell{30.8}{3.8}{42.3} & \heatcell{0.0}{0.0}{37.5} \\
Gemini-2.5-Flash~\protect\cite{gemini25flash} & -- & 1 FPS & \heatcell{27.4}{18.2}{41.1} & \heatcell{17.2}{17.0}{38.9} & \heatcell{30.4}{8.0}{40.2} & \heatcellb{70.0}{11.7}{70.0} & \heatcell{20.5}{10.3}{59.0} & \heatcell{0.0}{0.0}{61.5} & \heatcellb{31.9}{8.5}{31.9} & \heatcell{8.5}{0.0}{55.3} & \heatcell{0.0}{0.0}{48.9} & \heatcell{49.8}{12.3}{62.6} & \heatcell{11.5}{11.5}{61.5} & \heatcell{38.5}{19.2}{69.2} & \heatcellb{42.3}{3.8}{42.3} & \heatcell{31.2}{0.0}{37.5} \\
Kimi-K2.5~\protect\cite{kimi25} & -- & 1 FPS & \heatcell{29.3}{18.2}{41.1} & \heatcell{21.9}{17.0}{38.9} & \heatcell{17.9}{8.0}{40.2} & \heatcell{50.0}{11.7}{70.0} & \heatcell{38.5}{10.3}{59.0} & \heatcell{10.3}{0.0}{61.5} & \heatcell{10.6}{8.5}{31.9} & \heatcell{8.5}{0.0}{55.3} & \heatcell{19.1}{0.0}{48.9} & \heatcell{54.9}{12.3}{62.6} & \heatcell{30.8}{11.5}{61.5} & \heatcell{50.0}{19.2}{69.2} & \heatcellb{42.3}{3.8}{42.3} & \heatcell{18.8}{0.0}{37.5} \\
Qwen3.5-Plus~\protect\cite{qwen35suite} & -- & 1 FPS & \heatcell{28.1}{18.2}{41.1} & \heatcellb{38.9}{17.0}{38.9} & \heatcell{8.0}{8.0}{40.2} & \heatcell{60.0}{11.7}{70.0} & \heatcell{10.3}{10.3}{59.0} & \heatcell{30.8}{0.0}{61.5} & \heatcell{8.5}{8.5}{31.9} & \heatcell{0.0}{0.0}{55.3} & \heatcell{19.1}{0.0}{48.9} & \heatcell{16.2}{12.3}{62.6} & \heatcell{26.9}{11.5}{61.5} & \heatcellb{69.2}{19.2}{69.2} & \heatcell{30.8}{3.8}{42.3} & \heatcell{18.8}{0.0}{37.5} \\
Qwen3.6-Plus~\protect\cite{qwen36plus} & -- & 1 FPS & \heatcell{25.8}{18.2}{41.1} & \heatcell{24.0}{17.0}{38.9} & \heatcellb{40.2}{8.0}{40.2} & \heatcell{51.7}{11.7}{70.0} & \heatcell{10.3}{10.3}{59.0} & \heatcell{20.5}{0.0}{61.5} & \heatcell{10.6}{8.5}{31.9} & \heatcell{21.3}{0.0}{55.3} & \heatcell{31.9}{0.0}{48.9} & \heatcell{21.3}{12.3}{62.6} & \heatcell{19.2}{11.5}{61.5} & \heatcell{57.7}{19.2}{69.2} & \heatcellb{42.3}{3.8}{42.3} & \heatcell{0.0}{0.0}{37.5} \\
Qwen3.5-Flash~\protect\cite{qwen35suite} & -- & 1 FPS & \heatcell{25.9}{18.2}{41.1} & \heatcell{23.4}{17.0}{38.9} & \heatcell{11.6}{8.0}{40.2} & \heatcell{40.0}{11.7}{70.0} & \heatcell{30.8}{10.3}{59.0} & \heatcell{20.5}{0.0}{61.5} & \heatcell{10.6}{8.5}{31.9} & \heatcell{10.6}{0.0}{55.3} & \heatcell{0.0}{0.0}{48.9} & \heatcell{41.3}{12.3}{62.6} & \heatcell{30.8}{11.5}{61.5} & \heatcell{57.7}{19.2}{69.2} & \heatcell{30.8}{3.8}{42.3} & \heatcell{28.1}{0.0}{37.5} \\
\midrule
\rowcolor[HTML]{CBF1F5}
\multicolumn{17}{c}{\bfseries\itshape Open-Source MLLMs} \\
\midrule
Qwen3-VL-4B~\protect\cite{qwen3vl} & 4B & 1~FPS & \heatcell{23.1}{18.2}{41.1} & \heatcell{21.5}{17.0}{38.9} & \heatcell{35.7}{8.0}{40.2} & \heatcell{20.0}{11.7}{70.0} & \heatcell{28.2}{10.3}{59.0} & \heatcell{5.1}{0.0}{61.5} & \heatcell{12.8}{8.5}{31.9} & \heatcell{10.6}{0.0}{55.3} & \heatcell{23.4}{0.0}{48.9} & \heatcell{22.1}{12.3}{62.6} & \heatcell{38.5}{11.5}{61.5} & \heatcell{65.4}{19.2}{69.2} & \heatcell{11.5}{3.8}{42.3} & \heatcell{28.1}{0.0}{37.5} \\
MiniCPM-V 2.6~\protect\cite{minicpmv26} & 8B & 64 & \heatcell{23.1}{18.2}{41.1} & \heatcell{24.6}{17.0}{38.9} & \heatcell{17.0}{8.0}{40.2} & \heatcell{26.7}{11.7}{70.0} & \heatcell{28.2}{10.3}{59.0} & \heatcell{5.1}{0.0}{61.5} & \heatcell{17.0}{8.5}{31.9} & \heatcell{34.0}{0.0}{55.3} & \heatcell{25.5}{0.0}{48.9} & \heatcell{18.3}{12.3}{62.6} & \heatcell{42.3}{11.5}{61.5} & \heatcell{65.4}{19.2}{69.2} & \heatcell{19.2}{3.8}{42.3} & \heatcell{6.2}{0.0}{37.5} \\
Qwen3-VL-8B~\protect\cite{qwen3vl} & 8B & 1~FPS & \heatcell{26.2}{18.2}{41.1} & \heatcell{23.6}{17.0}{38.9} & \heatcell{30.4}{8.0}{40.2} & \heatcell{25.0}{11.7}{70.0} & \heatcell{35.9}{10.3}{59.0} & \heatcell{5.1}{0.0}{61.5} & \heatcell{25.5}{8.5}{31.9} & \heatcell{34.0}{0.0}{55.3} & \heatcell{21.3}{0.0}{48.9} & \heatcell{29.4}{12.3}{62.6} & \heatcell{34.6}{11.5}{61.5} & \heatcell{65.4}{19.2}{69.2} & \heatcell{3.8}{3.8}{42.3} & \heatcell{21.9}{0.0}{37.5} \\
InternVL3.5-8B~\protect\cite{internvl3.5} & 8B & 64 & \heatcell{27.0}{18.2}{41.1} & \heatcell{24.4}{17.0}{38.9} & \heatcell{29.5}{8.0}{40.2} & \heatcell{25.0}{11.7}{70.0} & \heatcell{38.5}{10.3}{59.0} & \heatcell{12.8}{0.0}{61.5} & \heatcell{17.0}{8.5}{31.9} & \heatcell{34.0}{0.0}{55.3} & \heatcell{17.0}{0.0}{48.9} & \heatcell{30.2}{12.3}{62.6} & \heatcell{38.5}{11.5}{61.5} & \heatcell{65.4}{19.2}{69.2} & \heatcell{19.2}{3.8}{42.3} & \heatcell{28.1}{0.0}{37.5} \\
LLaVA-Video-7B~\protect\cite{zhang2024llavanextvideo} & 7B & 1~FPS & \heatcell{24.9}{18.2}{41.1} & \heatcell{23.4}{17.0}{38.9} & \heatcell{27.7}{8.0}{40.2} & \heatcell{46.7}{11.7}{70.0} & \heatcell{38.5}{10.3}{59.0} & \heatcell{0.0}{0.0}{61.5} & \heatcell{14.9}{8.5}{31.9} & \heatcell{17.0}{0.0}{55.3} & \heatcell{2.1}{0.0}{48.9} & \heatcell{31.1}{12.3}{62.6} & \heatcell{26.9}{11.5}{61.5} & \heatcell{50.0}{19.2}{69.2} & \heatcell{11.5}{3.8}{42.3} & \heatcell{15.6}{0.0}{37.5} \\
LLaVA-Video-72B~\protect\cite{zhang2024llavanextvideo} & 72B & 1~FPS & \heatcell{29.4}{18.2}{41.1} & \heatcell{28.0}{17.0}{38.9} & \heatcell{32.1}{8.0}{40.2} & \heatcell{51.7}{11.7}{70.0} & \heatcell{41.0}{10.3}{59.0} & \heatcell{2.6}{0.0}{61.5} & \heatcell{19.1}{8.5}{31.9} & \heatcell{21.3}{0.0}{55.3} & \heatcell{6.4}{0.0}{48.9} & \heatcell{35.7}{12.3}{62.6} & \heatcell{30.8}{11.5}{61.5} & \heatcell{53.8}{19.2}{69.2} & \heatcell{15.4}{3.8}{42.3} & \heatcell{21.9}{0.0}{37.5} \\

\midrule
\rowcolor[HTML]{B9E5E8}
\multicolumn{17}{c}{\bfseries\itshape Specialized long-video methods} \\
\midrule
AKS~\protect\cite{AKS} & 8B & 128 & \heatcell{27.3}{18.2}{41.1} & \heatcell{25.0}{17.0}{38.9} & \heatcell{31.2}{8.0}{40.2} & \heatcell{30.0}{11.7}{70.0} & \heatcell{35.9}{10.3}{59.0} & \heatcell{2.6}{0.0}{61.5} & \heatcell{29.8}{8.5}{31.9} & \heatcell{23.4}{0.0}{55.3} & \heatcell{31.9}{0.0}{48.9} & \heatcell{34.0}{12.3}{62.6} & \heatcell{15.4}{11.5}{61.5} & \heatcell{50.0}{19.2}{69.2} & \heatcell{7.7}{3.8}{42.3} & \heatcell{18.8}{0.0}{37.5} \\
ViLAMP~\protect\cite{vilamp} & 8B & 128 & \heatcell{27.0}{18.2}{41.1} & \heatcell{24.8}{17.0}{38.9} & \heatcell{30.4}{8.0}{40.2} & \heatcell{21.7}{11.7}{70.0} & \heatcell{38.5}{10.3}{59.0} & \heatcell{5.1}{0.0}{61.5} & \heatcellb{31.9}{8.5}{31.9} & \heatcell{25.5}{0.0}{55.3} & \heatcell{27.7}{0.0}{48.9} & \heatcell{35.3}{12.3}{62.6} & \heatcell{15.4}{11.5}{61.5} & \heatcell{42.3}{19.2}{69.2} & \heatcell{7.7}{3.8}{42.3} & \heatcell{18.8}{0.0}{37.5} \\
WFS-SB~\protect\cite{chen2026wavelet} & 8B & 128 & \heatcell{28.1}{18.2}{41.1} & \heatcell{26.5}{17.0}{38.9} & \heatcell{31.2}{8.0}{40.2} & \heatcell{21.7}{11.7}{70.0} & \heatcell{38.5}{10.3}{59.0} & \heatcell{5.1}{0.0}{61.5} & \heatcellb{31.9}{8.5}{31.9} & \heatcell{27.7}{0.0}{55.3} & \heatcell{27.7}{0.0}{48.9} & \heatcell{36.6}{12.3}{62.6} & \heatcell{15.4}{11.5}{61.5} & \heatcell{42.3}{19.2}{69.2} & \heatcell{7.7}{3.8}{42.3} & \heatcell{18.8}{0.0}{37.5} \\
VideoLLaMA3-7B~\protect\cite{damonlpsg2025videollama3} & 7B & 1~FPS & \heatcell{22.7}{18.2}{41.1} & \heatcell{25.1}{17.0}{38.9} & \heatcell{18.8}{8.0}{40.2} & \heatcell{36.7}{11.7}{70.0} & \heatcell{20.5}{10.3}{59.0} & \heatcell{7.7}{0.0}{61.5} & \heatcell{29.8}{8.5}{31.9} & \heatcell{8.5}{0.0}{55.3} & \heatcell{10.6}{0.0}{48.9} & \heatcell{22.6}{12.3}{62.6} & \heatcell{15.4}{11.5}{61.5} & \heatcell{26.9}{19.2}{69.2} & \heatcell{7.7}{3.8}{42.3} & \heatcell{34.4}{0.0}{37.5} \\
LongVA-7B-DPO~\protect\cite{zhang2024longva} & 7B & 512 & \heatcell{18.2}{18.2}{41.1} & \heatcell{17.0}{17.0}{38.9} & \heatcell{20.5}{8.0}{40.2} & \heatcell{15.0}{11.7}{70.0} & \heatcell{33.3}{10.3}{59.0} & \heatcell{33.3}{0.0}{61.5} & \heatcell{21.3}{8.5}{31.9} & \heatcell{21.3}{0.0}{55.3} & \heatcell{14.9}{0.0}{48.9} & \heatcell{16.2}{12.3}{62.6} & \heatcell{23.1}{11.5}{61.5} & \heatcell{19.2}{19.2}{69.2} & \heatcell{3.8}{3.8}{42.3} & \heatcell{15.6}{0.0}{37.5} \\
VideoChat-Flash-7B~\protect\cite{li2024videochatflash} & 7B & 1~FPS & \heatcell{24.9}{18.2}{41.1} & \heatcell{25.1}{17.0}{38.9} & \heatcell{30.4}{8.0}{40.2} & \heatcell{11.7}{11.7}{70.0} & \heatcell{33.3}{10.3}{59.0} & \heatcell{0.0}{0.0}{61.5} & \heatcell{23.4}{8.5}{31.9} & \heatcell{10.6}{0.0}{55.3} & \heatcell{19.1}{0.0}{48.9} & \heatcell{31.1}{12.3}{62.6} & \heatcell{19.2}{11.5}{61.5} & \heatcell{61.5}{19.2}{69.2} & \heatcell{7.7}{3.8}{42.3} & \heatcell{21.9}{0.0}{37.5} \\
\midrule
\rowcolor[HTML]{A6E3E9}
\multicolumn{17}{c}{\bfseries\itshape Medical MLLMs} \\
\midrule
Lingshu-7B~\protect\cite{xu2025lingshu} & 7B & 1~FPS & \heatcell{25.0}{18.2}{41.1} & \heatcell{26.7}{17.0}{38.9} & \heatcell{23.2}{8.0}{40.2} & \heatcell{28.3}{11.7}{70.0} & \heatcell{46.2}{10.3}{59.0} & \heatcell{7.7}{0.0}{61.5} & \heatcell{12.8}{8.5}{31.9} & \heatcell{10.6}{0.0}{55.3} & \heatcell{19.1}{0.0}{48.9} & \heatcell{25.1}{12.3}{62.6} & \heatcell{34.6}{11.5}{61.5} & \heatcell{65.4}{19.2}{69.2} & \heatcell{19.2}{3.8}{42.3} & \heatcell{3.1}{0.0}{37.5} \\
Lingshu-32B~\protect\cite{xu2025lingshu} & 32B & 1~FPS & \heatcell{27.9}{18.2}{41.1} & \heatcell{24.0}{17.0}{38.9} & \heatcell{33.0}{8.0}{40.2} & \heatcell{38.3}{11.7}{70.0} & \heatcell{35.9}{10.3}{59.0} & \heatcell{12.8}{0.0}{61.5} & \heatcellb{31.9}{8.5}{31.9} & \heatcell{31.9}{0.0}{55.3} & \heatcell{17.0}{0.0}{48.9} & \heatcell{30.2}{12.3}{62.6} & \heatcell{50.0}{11.5}{61.5} & \heatcell{34.6}{19.2}{69.2} & \heatcell{15.4}{3.8}{42.3} & \heatcellb{37.5}{0.0}{37.5} \\
Hulu-Med-7B~\protect\cite{hulumed} & 7B & 1~FPS & \heatcell{26.2}{18.2}{41.1} & \heatcell{25.7}{17.0}{38.9} & \heatcell{17.0}{8.0}{40.2} & \heatcell{26.7}{11.7}{70.0} & \heatcell{38.5}{10.3}{59.0} & \heatcell{20.5}{0.0}{61.5} & \heatcell{25.5}{8.5}{31.9} & \heatcellb{55.3}{0.0}{55.3} & \heatcell{25.5}{0.0}{48.9} & \heatcell{23.8}{12.3}{62.6} & \heatcell{30.8}{11.5}{61.5} & \heatcell{34.6}{19.2}{69.2} & \heatcell{23.1}{3.8}{42.3} & \heatcell{25.0}{0.0}{37.5} \\
HuatuoGPT-Vision-7B~\protect\cite{huatuogptvision} & 7B & 1~FPS & \heatcell{20.1}{18.2}{41.1} & \heatcell{21.7}{17.0}{38.9} & \heatcell{23.2}{8.0}{40.2} & \heatcell{21.7}{11.7}{70.0} & \heatcell{53.8}{10.3}{59.0} & \heatcell{7.7}{0.0}{61.5} & \heatcell{8.5}{8.5}{31.9} & \heatcell{0.0}{0.0}{55.3} & \heatcell{34.0}{0.0}{48.9} & \heatcell{12.8}{12.3}{62.6} & \heatcell{30.8}{11.5}{61.5} & \heatcell{34.6}{19.2}{69.2} & \heatcell{19.2}{3.8}{42.3} & \heatcell{15.6}{0.0}{37.5} \\
MedGRPO~\protect\cite{medgrpo} & 7B & 1~FPS & \heatcell{20.1}{18.2}{41.1} & \heatcell{22.1}{17.0}{38.9} & \heatcell{19.6}{8.0}{40.2} & \heatcell{23.3}{11.7}{70.0} & \heatcell{35.9}{10.3}{59.0} & \heatcell{12.8}{0.0}{61.5} & \heatcell{19.1}{8.5}{31.9} & \heatcell{8.5}{0.0}{55.3} & \heatcellb{48.9}{0.0}{48.9} & \heatcell{12.3}{12.3}{62.6} & \heatcell{26.9}{11.5}{61.5} & \heatcell{19.2}{19.2}{69.2} & \heatcell{11.5}{3.8}{42.3} & \heatcell{9.4}{0.0}{37.5} \\

\bottomrule
\end{tabular}
\end{adjustbox}

\end{table*}




Table~\ref{tab:main_results} shows that long medical video understanding remains substantially unsolved. The strongest model, Gemini-3.1-Pro-Preview, reaches only 41.1\% overall accuracy, while most open-source and medical MLLMs remain around the mid-20s. This gap is notable because evaluation is conducted under full-procedure video context. Models must therefore not only recognize medical visual content, but also search long redundant streams for sparse evidence and use it for clinical judgment. The per-task results reveal this bottleneck more clearly than the overall score. Tasks that can benefit from broader procedural regularities or salient visual cues, such as BPQ, RL, IR, and WR, show relatively higher peaks. In contrast, lesion-centric tasks such as LLS, Site, Hist, and Size are more unstable across models, since they require locating subtle local findings and binding them to anatomical sites or clinical attributes. Semantic-reasoning tasks further expose the weakness of current long-video reasoning. RSR is often stronger, suggesting that procedural order provides a usable cue, whereas CPR and LA remain difficult because they require counting, comparing, and aggregating distributed observations across the same procedure.

These results underscore the core finding of \ours: the primary obstacle is the coupling between sparse evidence localization and cross-temporal reasoning. Although current MLLMs exploit coarse global cues or isolated salient frames, they struggle when answers depend on subtle evidence, attribute binding, or aggregating distant events. This suggests longer context windows are insufficient without mechanisms to selectively retrieve weak evidence and maintain structured clinical context throughout the procedure.

\begin{wrapfigure}{r}{0.45\linewidth}

  \centering
  \includegraphics[width=\linewidth]{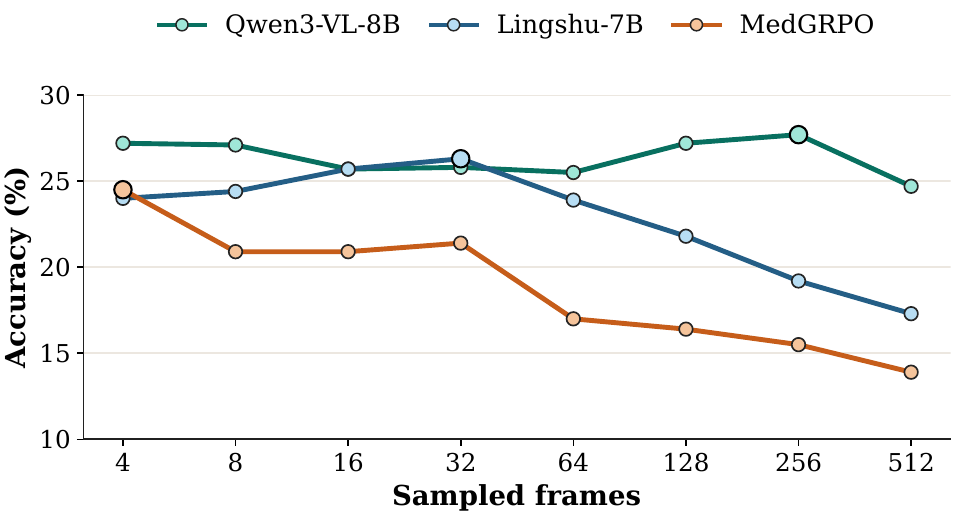}
  \caption{Frame-scaling performance under different sampled-frame budgets.}
  \label{fig:frame_scaling}

\end{wrapfigure}

\subsection{RQ1: Frame Scaling Is Unreliable in Medical Long Videos.}



Figure~\ref{fig:frame_scaling} tests whether medical long-video accuracy improves as more frames are sampled from the same videos. The main observation is the absence of a consistent positive slope. Qwen3-VL-8B is comparatively robust, but its curve mostly oscillates within a narrow band: it improves again at 128--256 frames and then drops at 512 frames. Lingshu-7B benefits only from a small-to-moderate budget, reaching its best result at 32 frames before steadily declining as the input becomes longer. MedGRPO shows the clearest context-dilution effect: adding frames does not recover more evidence, but instead drives accuracy downward from 4 to 512. 

These different curve shapes indicate that frame count is not the controlling variable for reliable medical video understanding. More frames can improve the chance of including sparse clinical evidence, but they also add near-duplicate anatomy, low-information procedural segments, and visually similar distractors. Current models do not consistently convert the extra visual context into better evidence use: Qwen3-VL-8B tolerates longer inputs without a clear gain, while the medical-domain models degrade once redundant context dominates the useful signal. Thus, the bottleneck is not merely insufficient temporal coverage, but the lack of evidence-aware selection and compression. Long-context medical MLLMs need mechanisms that retrieve subtle decisive frames and suppress redundant context, rather than simply increasing the sampled-frame budget.


\begin{figure}[t]
    \centering
    \includegraphics[width=0.8\linewidth]{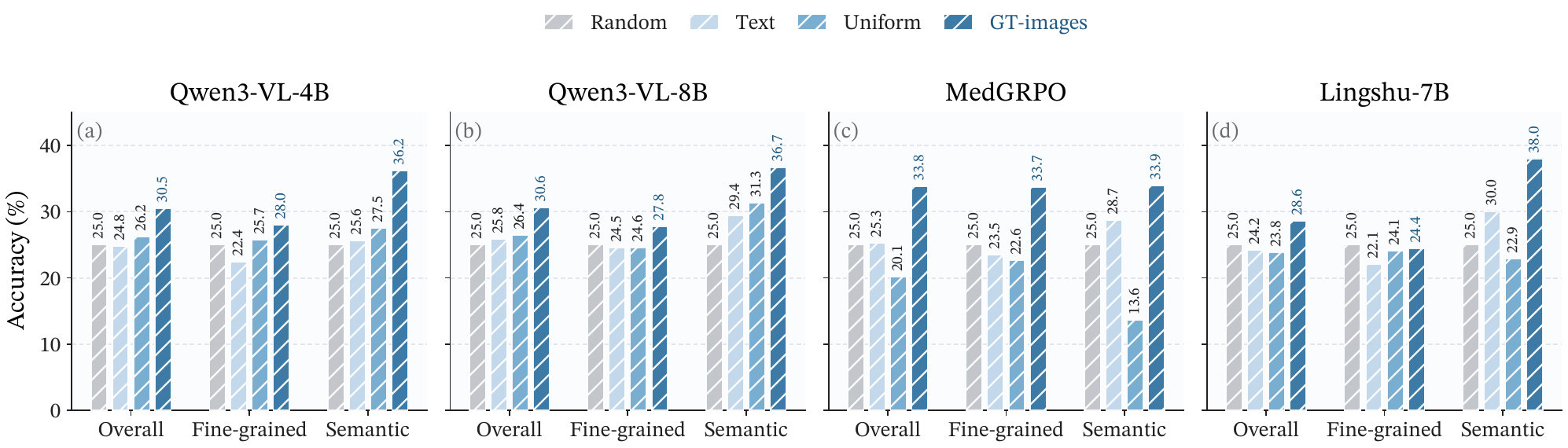}
    \caption{Diagnostic breakdown under four input settings: random guessing, text-only questions, uniformly sampled full-video frames, and oracle GT-images. The comparison separates language shortcuts, temporal evidence retrieval, and post-retrieval evidence interpretation.}
    \label{fig:e7}
\end{figure}

\subsection{RQ2: Evidence Retrieval and Interpretation Remain Bottlenecks.}

Figure~\ref{fig:e7} diagnoses evidence utilization across four settings: \textit{Random} (25\% baseline), \textit{Text} (no visual input), \textit{Uniform} (full-procedure sampling), and \textit{GT-images} (human-verified oracle frames). This separates language shortcuts, temporal retrieval, and visual interpretation. Results show \ours is not solvable by language priors alone; text-only accuracy mirrors the random baseline, while uniform video input yields limited, unstable gains. Notably, MedGRPO drops to 13.6\% on Semantic tasks under uniform sampling, suggesting irrelevant frames can mislead understanding when sparse evidence is missed. The Uniform-to-GT-images gap highlights a retrieval bottleneck, particularly for Semantic reasoning: MedGRPO improves from 13.6\% to 33.9\%, and Lingshu-7B from 22.9\% to 38.0\%. However, even with oracle evidence, all models remain below 40\%, proving retrieval is only the initial failure mode. Models must still identify subtle findings, distinguish similar anatomy, and bind observations to phases or events. Thus, RQ2 exposes a dual bottleneck: MLLMs require both evidence-aware temporal retrieval and clinically grounded visual verification.

\begin{figure}[t]
    \centering
    \includegraphics[width=0.75\linewidth]{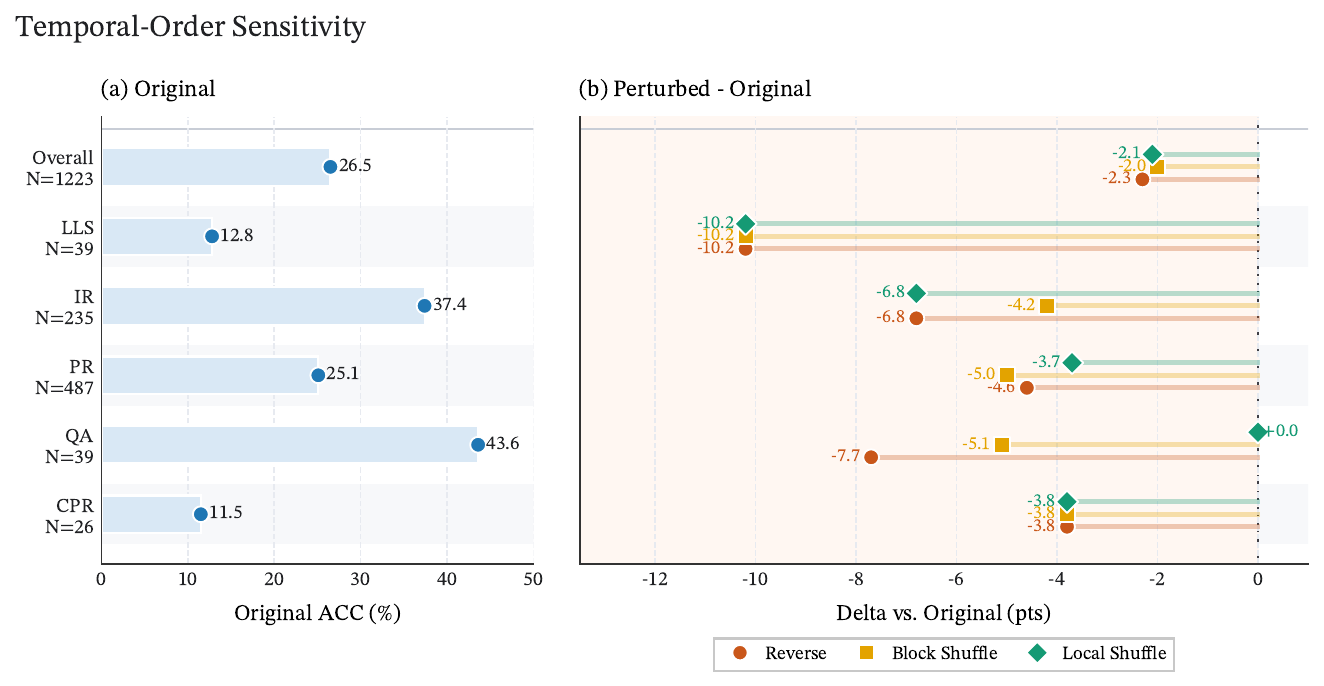}
    \caption{Disrupting event order consistently hurts performance. That there is only a slight performance decrease underscores the models' inability to effectively exploit temporal information.}
    \label{fig:e1}
\end{figure}

\begin{figure}[t]
    \centering

        \includegraphics[width=0.9\linewidth]{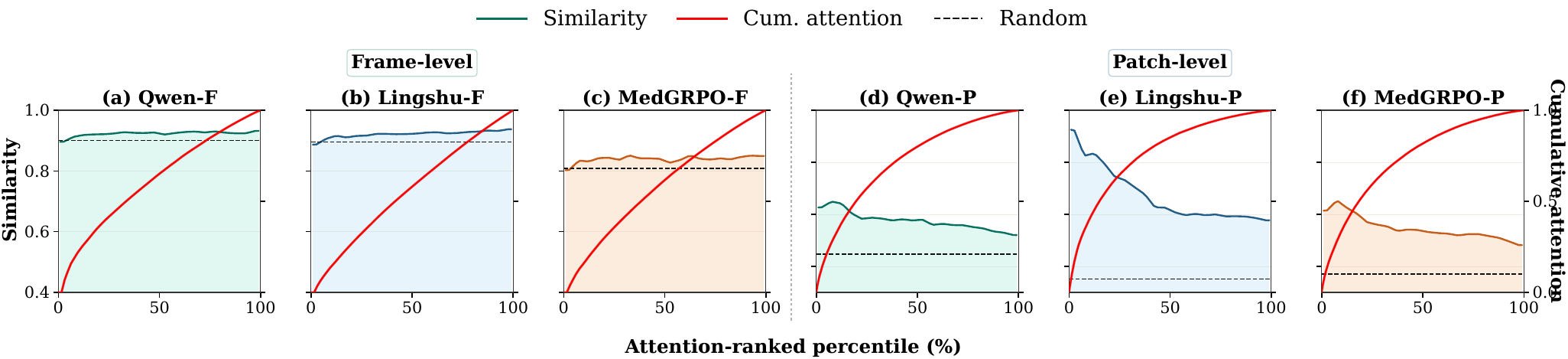}
    \caption{Attention distribution and similarity in frame and patch-level.}
    \label{fig:e6}
\end{figure}

\begin{figure}[t]
    \centering
    \includegraphics[width=0.8\linewidth]{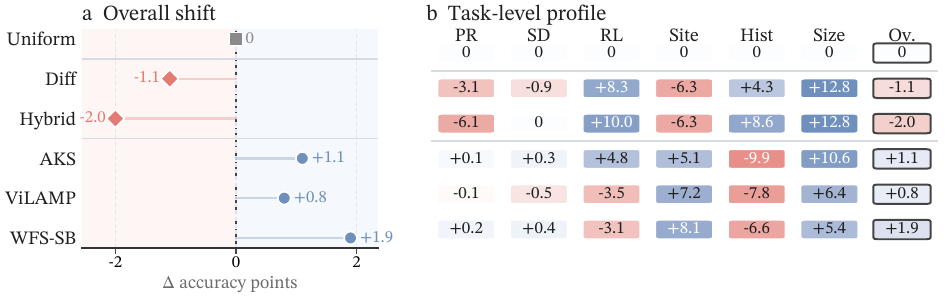}
    \caption{Frame-change sampling helps short-window tasks, but uniform sampling remains the most stable overall strategy. And other generic sampling methods also have limited improvement.}
    \label{fig:e2}
\end{figure}


\subsection{RQ3: Weak Procedural Reasoning and Attention Drift under Redundancy.}

  We analyze the mechanisms behind these bottlenecks by diagnosing how models handle temporal
  structure and visual attention distribution. Figure~\ref{fig:e1} perturbs frame order to measure
  whether models use procedural chronology, while Figure~\ref{fig:e6} follows ViLAMP~\cite{vilamp} to
  diagnose whether attention concentrates on informative frame- and patch-level evidence. For each
  video-question pair, we sample 128 frames, aggregate query-to-visual attention into frame scores,
  and plot cumulative attention after ranking frames by attention. For patch-level analysis, we use
  the top 32 attended frames as keyframes and compare patches in the remaining 96 frames with their
  corresponding patches in the nearest preceding keyframe. Thus, the left half of Figure~\ref{fig:e6}
  measures temporal redundancy, and the right half measures spatial redundancy. Figure~\ref{fig:e1} shows that current models use temporal order only weakly. Reversing, block-
  shuffling, or locally shuffling frames reduces Qwen3-VL-8B from 26.5\% to 24.2--24.5\% overall,
  with larger losses on workflow- or context-sensitive tasks such as PR, IR, QA, LLS, and CPR. This
  confirms that chronology matters, but the small overall drop suggests that the model mainly
  captures coarse temporal cues rather than robust procedural structure. Figure~\ref{fig:e6} further shows that medical videos break the general-video intuition that
  attention should collapse onto a few salient frames. The frame-level cumulative curves increase
  smoothly, so Qwen3-VL and Lingshu still require 76.7\% and 85.6\% of temporal units to accumulate
  90\% attention, and MedGRPO remains similarly diffuse. Meanwhile, Qwen3-VL-8B and Lingshu-7B
  maintain very high frame similarity around 0.90, while MedGRPO is slightly lower, suggesting that
  medical-video training may improve frame discriminability but not evidence-selective attention. The
  patch-level panels show the same issue spatially: high-attention patches remain highly similar to
  keyframe patches, especially for Lingshu-7B. We refer to this phenomenon as \textit{attention
  drift}: under strong spatiotemporal redundancy, model attention drifts toward visually repetitive
  but clinically non-decisive frames and patches, rather than reliably concentrating on sparse
  answer-relevant evidence. Thus, medical redundancy is deceptive: many frames and patches look
  alike, but the answer often depends on a subtle outlier that current attention patterns fail to
  isolate.

\subsection{RQ4: Generic Sampling Trades Local Detail for Global Coverage.}

Figure~\ref{fig:e2} evaluates whether sampling mitigates long-video challenges. Uniform sampling remains stable by preserving procedural coverage but is evidence-agnostic, often missing sparse findings. Conversely, change-focused samplers improve short-window tasks like RL and Size but hinder workflow-sensitive tasks (e.g., PR), showing that visual dynamism does not guarantee clinical relevance. Table~\ref{tab:main_results} confirms specialized sampling is promising: with 128 frames, AKS, ViLAMP, and WFS-SB achieve 27.3\%, 27.0\%, and 28.1\% accuracy, outperforming general MLLMs. Gains are localized; notably, WFS-SB reaches 32.8\% on Site. This reflects their designs: ViLAMP prioritizes keyframe fidelity via compression, while WFS-SB employs semantic-boundary selection to retain structured temporal context. These improvements remain modest and task-dependent. The methods still underperform on tasks requiring global procedural continuity or multi-step aggregation, such as WR and RSR, where aggressive keyframe selection may discard temporal context for reasoning. Sampling can reduce redundancy and recover some sparse evidence, but cannot solve long medical video understanding alone. Future methods need clinically evidence-aware sampling that preserves global procedural structure, retrieves weak local findings, and verifies anatomical and temporal context before reasoning.

\section{Conclusion}

We present \ours, an in-the-wild benchmark for long-context medical video understanding.  Representative general-domain, medical-domain, and retrieval-style MLLMs perform poorly, with the best model reaching only 41.1\% accuracy. Our analyses show that non-monotonic frame scaling, persistent bottlenecks in clinical evidence retrieval and interpretation, the weak procedural reasoning and attention drift, and the limited ability of generic sampling strategies. \ours remains limited by public-video diversity, annotation quality, and a multiple-choice protocol that cannot fully capture free-form clinical reporting, but offers a clear testbed for sparse evidence interpretation and full-procedure clinical reasoning in long-context medical videos.

\bibliographystyle{assets/plainnat}
\bibliography{main}

\clearpage
\appendix

\section{Benchmark Card and Construction Details}
\label{sec:supp_benchmark_card}

\paragraph{Purpose.}
\ours is an evaluation benchmark for full-procedure medical video understanding. The benchmark is designed to test whether an MLLM can search a long, redundant clinical video for sparse evidence and then answer a multiple-choice clinical question that may require local recognition, anatomical binding, temporal comparison, or procedure-level aggregation. The benchmark is not intended for autonomous diagnosis or deployment-time clinical decision making.

\paragraph{Final artifact.}
The released evaluation split contains 1,253 multiple-choice questions over 340 public or previously released clinical videos. Each item contains a source video pointer, a question, four answer choices, a single final answer label, task metadata, source provenance, and evidence metadata used for audit and oracle-evidence analyses. Table~\ref{tab:artifact_card} summarizes the final artifact.

\begin{table}[h]
\centering
\caption{Benchmark card summary for the final evaluation artifact.}
\label{tab:artifact_card}
\small
\begin{tabularx}{\linewidth}{P{0.32\linewidth}Y}
\toprule
\textbf{Field} & \textbf{Documentation} \\
\midrule
Benchmark size & 340 videos; 759 decoded hours; 1,253 QA items. \\
Clinical scope & Surgery, colonoscopy, capsule/endoscopic examination, and ultrasound; 7 organ or anatomy groups. \\
Question format & Four-way multiple choice. The prompt asks the model to return only A/B/C/D. \\
Task families & Fine-grained understanding: PR, SD, RL, BPQ, LLS, Site, Hist, Size. Semantic reasoning: IR, WR, RSR, CPR, LA. \\
Evidence grounding & Every retained item is linked to source annotations and reviewed for video usability, QA well-formedness, evidence support, and final-label consistency. \\
Release mode & Derived benchmark metadata, prompts, labels, evaluation scripts, and evidence metadata are released. Raw video redistribution follows the upstream dataset license; where redistribution is not permitted, we release source pointers and reconstruction instructions instead of re-hosting the video. \\
Responsible use & The benchmark is for research evaluation. It should not be interpreted as validating a model for clinical deployment. \\
\bottomrule
\end{tabularx}
\end{table}

\subsection{Source Data, Provenance, and Release Handling}
\label{sec:supp_source_data}

Table~\ref{tab:source_documentation} expands the source table in the main paper with provenance, annotation use, and release handling.There are 8 origin datasets in total: AutoLaparo  \cite{wang2022autolaparonewdatasetintegrated}, Galar \cite{le2025galar}, GraSP \cite{ayobi2025pixelgrasp}, MultiBypass  \cite{lavanchy2024challengesutlibypass}, PitVQA \cite{he2024pitvqa}, REAL-Colon \cite{biffi2024realcolon}, TMVP \cite{lan2025new} and US-Study \cite{farras2024realusstudy}. We preserve the original source identity for every item. We do not convert upstream clinical video licenses into a new raw-video license; users must follow each original dataset's terms. The derived QA/evidence metadata are released separately under the benchmark release terms.

\begin{table*}[t]
\centering
\caption{Source-dataset documentation. ``Release handling'' describes how the benchmark package respects upstream asset terms.}
\label{tab:source_documentation}
\scriptsize
\begin{adjustbox}{width=\linewidth}
\begin{tabular}{l r r r P{0.24\linewidth} P{0.27\linewidth} P{0.25\linewidth}}
\toprule
\textbf{Dataset} & \textbf{Videos} & \textbf{QA} & \textbf{Hours} & \textbf{Annotation source used} & \textbf{Evidence extraction rule} & \textbf{License / release handling} \\
\midrule
AutoLaparo   & 20 & 44 & 22.3 & Public laparoscopic phase and workflow labels. & Time-window questions are aligned to phase labels and nearby annotated segments. & Public research asset; raw-video access follows upstream terms. We release derived QA/evidence metadata and source pointers. \\
Galar & 44 & 60 & 351.3 & Public GI-tract region labels and temporal section information. & Region-localization evidence is the annotated temporal section or the queried time window. & Public research asset; raw-video redistribution is not assumed. We release source pointers and derived metadata. \\
GraSP  & 13 & 112 & 32.4 & Public prostatectomy workflow and instrument annotations. & Phase, step, instrument, workload, ranking, and counting questions use the corresponding temporal annotation tracks. & Public research asset; users must comply with upstream dataset terms. \\
MultiBypass  & 125 & 403 & 193.3 & Public bariatric surgery phase/step annotations. & Phase and step questions use exact annotated temporal intervals; distractors are sampled from plausible nearby labels. & Public research asset; derived metadata are released separately from raw videos when needed. \\
PitVQA & 10 & 13 & 10.5 & Public pituitary surgery QA / visual annotation source. & Questions are retained only when the corresponding video and label are usable under the long-video protocol. & Public research asset; original citation and access path are preserved. \\
REAL-Colon  & 39 & 251 & 17.0 & Public colonoscopy findings, lesion attributes, and bowel-preparation labels. & Lesion-level evidence is the union of annotated lesion frames or lesion proxy windows; exam-level quality uses the full colonoscopy as support. & Public clinical-video asset; raw redistribution follows upstream terms. Derived lesion-QA metadata are released with provenance fields. \\
TMVP  & 59 & 340 & 122.7 & Public cardiac surgery phase and instrument annotations. & Phase questions use exact phase intervals; instrument-onset questions use a short window around the annotated phase transition. & Public research asset; source identity and annotation provenance are preserved. \\
US-Study  & 30 & 30 & 9.6 & Public ultrasound study metadata and full-study labels. & Study-level contraction questions use the full examination because the public metadata are study-level rather than frame-level. & Public study source; we release derived QA metadata and source pointers. \\
\midrule
Total & 340 & 1,253 & 759.0 & -- & -- & -- \\
\bottomrule
\end{tabular}
\end{adjustbox}
\end{table*}

\subsection{Question Construction Pipeline}
\label{sec:supp_construction_pipeline}

The construction pipeline is evidence-first rather than template-first. For each candidate item, we begin with a source annotation event, window, lesion record, or study label. We then generate a candidate question whose answer can be traced back to that source evidence. Distractors are drawn from clinically plausible labels in the same task family, with preference for temporally nearby or visually confusable alternatives. The question is rewritten into direct natural language while preserving the option set and answer label. Finally, the item is reviewed by humans before entering the final evaluation file.

The final construction pipeline is:
\begin{enumerate}
    \item \textbf{Source normalization.} Decode videos, standardize source identifiers, remove unusable/corrupt videos, and retain the original source annotation fields needed for provenance.
    \item \textbf{Evidence retrieval.} Resolve a source annotation into an evidence unit: an exact frame window, a short onset window, a lesion-support union, or a full-study support tag.
    \item \textbf{Candidate QA generation.} Convert the evidence unit into a task-specific multiple-choice question and construct plausible distractors from the same label space.
    \item \textbf{Automatic filtering.} Remove temporally invalid, ambiguous, duplicate, or malformed candidates; balance answer letters by option shuffling.
    \item \textbf{Language rewriting.} Rewrite questions into natural standalone prompts while preserving the answer label and option semantics.
    \item \textbf{Human verification.} Three reviewers audit every final retained item for video usability, QA well-formedness, evidence support, and label consistency.
\end{enumerate}

\paragraph{Task template forms.}
Task templates are evidence-binding schemas rather than free-form question generators. Each template specifies the source annotation type, the queried clinical variable, the answer label space, the distractor pool, and the evidence-support rule. The main template families are: temporal-window recognition for phase, step, or anatomical-region labels; phase-onset instrument recognition; lesion identity to site, histology, or size binding; procedure-level counting, ranking, and proportion reasoning; and exam/study-level assessment such as bowel-preparation quality or ultrasound contraction recognition.

\paragraph{Automatic filtering rules.}
Automatic validity filtering removes a candidate if the source video is missing or not decodable, the evidence window falls outside the decoded video duration, the queried temporal window crosses incompatible source labels without being a global-reasoning task, the answer label is missing from the original annotation, the four options are not unique, more than one option can be correct under the source annotation, or the question becomes malformed after path/label normalization. The final hard-set curation removes 81 invalid candidates: 41 temporal-overrun questions and 40 ambiguous phase-instrument questions.

\paragraph{Distractor strengthening.}
Distractors are chosen from the same task label space as the answer and are rejected if they are also supported by the selected evidence. For temporal phase/step questions, distractors prefer nearby or visually confusable phases from the same procedure family. For instrument and lesion tasks, distractors come from plausible instruments or lesion attributes observed in the same source domain when possible. For counting, ranking, and proportion questions, distractors are adjacent bins or alternative phases that remain clinically plausible. Final option shuffling is applied after strengthening so that the correct answer letter is balanced across the benchmark.

\paragraph{Final curation funnel.}
The final hard-candidate pool is processed jointly across all source datasets. It contains 1,334 source-annotation-derived QA candidates before filtering. Automatic validity filtering removes 81 invalid items, leaving 1,253 final QA items. The language-rewrite stage preserves all answer labels, changes 1,253/1,253 question texts in the final natural-language export, and cleans 132 option texts without changing the gold option. Human verification is applied after these transformations, and only items passing the final data-review checks are retained.

\subsection{Human Verification Protocol}
\label{sec:supp_human_review}

All 1,253 final retained items pass human review by three data reviewers. The review is an audit layer over existing source annotations rather than a new clinical diagnosis process. The reviewers are non-physician data reviewers trained on the benchmark schema and the corresponding source-task metadata. This role is appropriate because they do not assign new medical diagnoses or overwrite clinical findings; instead, they calibrate each constructed QA item against the professional annotations already provided by the original medical-video datasets.

For every retained item, reviewers inspect the video, question, options, evidence metadata, and source annotation. They check whether the video is decodable and usable for the intended task, whether the question and answer choices are well formed, whether the selected evidence supports the answer under the task-specific evidence rule, and whether the final label remains consistent with the original source annotation after option shuffling and rewriting. The review has no fixed per-item time budget: reviewers follow the source evidence and metadata until the item either passes the four checks or exposes a traceable issue. Items are revised when the correction is traceable to the source annotation; otherwise they are removed. The final released set contains no unresolved audit failures.

\paragraph{Reviewer qualification and adjudication.}
The audit uses the original dataset annotation as the authority for procedure-specific semantics. A reviewer checks whether the visual evidence and rewritten prompt preserve that annotation. Ambiguous or disputed items are adjudicated by a senior reviewer; unresolved cases are not retained. No patient interaction, prospective recruitment, or independent diagnostic labeling is involved. We therefore report the process as consensus data review against source annotations, not as independent medical voting.

\paragraph{Audit statistics.}
The final audit covers 1,253/1,253 items. The final answer distribution is balanced after option shuffling: A: 309, B: 303, C: 319, D: 322. The final benchmark contains 689 temporal-localization items and 564 compositional-reasoning items. By source dataset, the item counts are AutoLaparo 44, Galar 60, GraSP 112, MultiBypass 403, PitVQA 13, REAL-Colon 251, TMVP 340, and US-Study 30.

\paragraph{Task-count imbalance.}
Question counts differ across tasks because the upstream datasets expose different annotation types and levels of granularity. The final per-task counts are PR 517, SD 112, RL 60, BPQ 39, LLS 39, Site 47, Hist 47, Size 47, IR 235, WR 26, RSR 26, CPR 26, and LA 32. Main-table task accuracies are computed as item-level correct/total ratios with these denominators and rounded to one decimal place. We therefore interpret small-task accuracies as directional diagnostics rather than highly stable standalone rankings. We do not synthesize additional items for under-represented clinical tasks because \ours requires every question to be grounded in original professional annotations and auditable video evidence.

\subsection{Operational Definition of Evidence Sparsity}
\label{sec:supp_evidence_sparsity}

Evidence frames are not independently relabeled medical frames. They are deterministically derived from upstream annotation timestamps, event intervals, lesion annotations, or study-level labels. Exact event annotations become frame windows; phase-onset questions use short windows around the annotated transition; lesion questions use the annotated lesion frame set or lesion-proxy union; study-level items are marked as global support and excluded from local evidence-sparsity estimation.

Evidence sparsity is computed at the question level. Let $V_q$ be the full source video for question $q$, $N_q$ be the number of decoded frames in $V_q$, and $E_q$ be the union of evidence frames needed to support the answer under the task-specific evidence rule. The question-level evidence ratio is
\begin{equation}
    r_q = \frac{|E_q|}{N_q}.
\end{equation}
The reported frame sparsity is the mean of $r_q$ over questions with localized evidence windows. Whole-study questions, such as study-level ultrasound or exam-level bowel-preparation quality, are treated as global-support questions and are not used to estimate local evidence sparsity, because their correct support is intentionally the full procedure rather than a sparse moment.

For the sparsity audit used in the main paper, the localizable subset contains 1,035 questions. Its mean evidence ratio is 0.166\%, equivalent to 1.7 evidence frames per 1,000 decoded frames, and its median ratio is 0.051\%. The audit export distinguishes exact local windows from proxy lesion unions: 894 questions use exact or directly localized support, and 141 use lesion-union proxy support for lesion-centric colonoscopy questions where the clinical source provides lesion-level annotations rather than a single contiguous answer clip.

\begin{table}[h]
\centering
\caption{Task-level evidence-support rules.}
\label{tab:evidence_rules}
\small
\begin{tabularx}{\linewidth}{P{0.20\linewidth} P{0.22\linewidth} Y}
\toprule
\textbf{Task family} & \textbf{Evidence unit} & \textbf{Rule} \\
\midrule
PR/SD/RL & Temporal window or annotated segment & Evidence is the annotated phase, step, or anatomical-region interval that overlaps the queried time window. \\
IR & Onset-centered short window & Evidence is a short window around the annotated phase transition where the queried instrument is visible. \\
WR/RSR/CPR & Procedure-level aggregation & Evidence consists of source annotation tracks over the full procedure; these are global-support reasoning questions rather than sparse-window questions. \\
BPQ/LLS/LA & Exam-level or lesion-union support & Global quality questions use full-exam support; lesion-size/distribution questions use the union of annotated lesion evidence. \\
Site/Hist/Size & Lesion-specific support & Evidence is the annotated lesion frame set or the proxy window associated with the lesion mentioned by the question. \\
US-Study & Full-study support & Evidence is the full ultrasound study because the released annotation is study-level. \\
\bottomrule
\end{tabularx}
\end{table}

\section{Main-Result Evaluation Protocol}
\label{sec:supp_main_protocol}

\subsection{Frame Admission and Cap Handling}
\label{sec:supp_frame_admission}

The main paper reports each model under the closest reproducible official or released inference setting. The key point for full-procedure validity is that capped inputs are never produced by prefix truncation. Whenever a model uses fps-based video input, we first decode the whole procedure at the requested fps and then uniformly subsample over the entire decoded sequence if the frame count exceeds the model-specific cap. Therefore, a 37.2-hour video is still represented across its full temporal extent, although sparsely, rather than only by the first segment.

The sampling rule is:
\begin{enumerate}
    \item Decode candidate frame indices at the model fps, normally 1 FPS.
    \item If the number of candidate indices is at most the cap $C$, keep all candidates.
    \item If the number of candidates exceeds $C$, keep $C$ indices uniformly spaced from the first to the last candidate.
    \item Preserve chronological order in the model input.
\end{enumerate}
For fixed-frame rows, we apply the same uniform-over-full-video rule directly to the full decoded frame index sequence. For specialized sampling methods, we follow the released method-specific selector and only constrain the final selected budget to the value reported in the Input column.

\subsection{Official Sources and Model-Specific Settings}
\label{sec:supp_model_settings}

Tables~\ref{tab:closed_protocol} and~\ref{tab:open_protocol} provide the protocol table corresponding to Table 2 in the main paper. The ``official source'' column gives the public model page, official repository, or official documentation used to align the inference pathway.

\begin{table*}[t]
\centering
\caption{Closed-source model protocol. All capped video inputs use uniform temporal subsampling over the full procedure, not prefix truncation.}
\label{tab:closed_protocol}
\scriptsize
\begin{adjustbox}{width=\linewidth}
\begin{tabular}{l P{0.30\linewidth} c c P{0.34\linewidth}}
\toprule
\textbf{Rows} & \textbf{Official source} & \textbf{Input shown} & \textbf{Cap} & \textbf{Protocol} \\
\midrule
GPT-5.4-mini; GPT-5.4 & \href{https://openai.com/index/introducing-gpt-5-4/}{OpenAI GPT-5.4}; \href{https://openai.com/index/introducing-gpt-5-4-mini-and-nano/}{OpenAI GPT-5.4 mini} & 1 FPS & 1,248 frames & Decode at 1 FPS; uniformly subsample to at most 1,248 frames over the full procedure; submit as native multimodal input. \\
Gemini-3.1-Pro-Preview; Gemini-2.5-Flash & \href{https://blog.google/innovation-and-ai/models-and-research/gemini-models/gemini-3-1-pro/}{Gemini 3.1 Pro}; \href{https://blog.google/products/gemini/gemini-2-5-model-family-expands}{Gemini 2.5 family} & 1 FPS & 1,248 frames & Same full-procedure 1 FPS sampling and uniform cap rule; no prefix truncation. \\
Kimi-K2.5 & \href{https://www.kimi.com/blog/kimi-k2-5}{Kimi K2.5} & 1 FPS & 1,248 frames & Same full-procedure 1 FPS sampling and uniform cap rule. \\
Qwen3.5-Plus; Qwen3.6-Plus; Qwen3.5-Flash & \href{https://help.aliyun.com/zh/model-studio/newly-released-models}{Alibaba Cloud Model Studio release notes} & 1 FPS & 1,248 frames & Same full-procedure 1 FPS sampling and uniform cap rule before API upload. \\
\bottomrule
\end{tabular}
\end{adjustbox}
\end{table*}

\begin{table*}[t]
\centering
\caption{Open-source, specialized long-video, and medical-model protocol. Fixed-frame rows use uniform sampling over the full source video unless the method is an adaptive selector.}
\label{tab:open_protocol}
\scriptsize
\begin{adjustbox}{width=\linewidth}
\begin{tabular}{l c c P{0.54\linewidth}}
\toprule
\textbf{Model / row} & \textbf{Input shown} & \textbf{Cap or budget} & \textbf{Official source} \\
\midrule
Qwen3-VL-4B; Qwen3-VL-8B & 1 FPS & official processor cap & \url{https://github.com/QwenLM/Qwen3-VL}; \url{https://huggingface.co/Qwen/Qwen3-VL-8B-Instruct} \\
MiniCPM-V 2.6 & 64 & 64 frames & \url{https://github.com/OpenBMB/MiniCPM-V}; \url{https://huggingface.co/openbmb/MiniCPM-V-2_6} \\
InternVL3.5-8B & 64 & 64 frames & \url{https://github.com/OpenGVLab/InternVL}; \url{https://huggingface.co/OpenGVLab/InternVL3_5-8B-Instruct} \\
LLaVA-Video-7B; LLaVA-Video-72B & 1 FPS & 64 frames & \url{https://github.com/LLaVA-VL/LLaVA-NeXT}; model cards under \url{https://huggingface.co/lmms-lab} \\
AKS & 128 & 128 selected frames & \url{https://github.com/ncTimTang/AKS} \\
ViLAMP & 128 & 128 selected frames & \url{https://github.com/steven-ccq/ViLAMP} \\
WFS-SB & 128 & 128 selected frames & \url{https://github.com/MAC-AutoML/WFS-SB} \\
VideoLLaMA3-7B & 1 FPS & 128 frames & \url{https://github.com/DAMO-NLP-SG/VideoLLaMA3}; \url{https://huggingface.co/DAMO-NLP-SG/VideoLLaMA3-7B} \\
LongVA-7B-DPO & 512 & 512 frames & \url{https://github.com/EvolvingLMMs-Lab/LongVA}; \url{https://huggingface.co/lmms-lab/LongVA-7B-DPO} \\
VideoChat-Flash-7B & 1 FPS & 512 frames & \url{https://github.com/OpenGVLab/VideoChat-Flash}; \url{https://huggingface.co/OpenGVLab/VideoChat-Flash-Qwen2_5-7B_InternVideo2-1B} \\
Lingshu-7B; Lingshu-32B & 1 FPS & official Qwen-style cap & \url{https://huggingface.co/lingshu-medical-mllm/Lingshu-7B}; \url{https://alibaba-damo-academy.github.io/lingshu/} \\
Hulu-Med-7B & 1 FPS & 1,800 frames & \url{https://github.com/ZJUI-AI4H/Hulu-Med}; \url{https://huggingface.co/ZJU-AI4H/Hulu-Med-7B} \\
HuatuoGPT-Vision-7B & 1 FPS & official Qwen-style cap & \url{https://github.com/FreedomIntelligence/HuatuoGPT-Vision}; \url{https://huggingface.co/FreedomIntelligence/HuatuoGPT-Vision-7B-Qwen2.5VL} \\
MedGRPO & 1 FPS & official released cap & \url{https://arxiv.org/abs/2512.06581}; project page: \url{https://uii-america.github.io/} \\
\bottomrule
\end{tabular}
\end{adjustbox}
\end{table*}

\paragraph{Why the main table is not a matched-budget leaderboard.}
The main table intentionally reports models under their closest official inference path because closed-source APIs, native long-video models, fixed-frame models, and adaptive samplers expose different public interfaces. As a result, the table should be read as an evaluation of deployable/released settings, not as a pure architecture-only ranking. The frame-scaling and sampling analyses in the main paper are included to separate model capability from frame-budget effects.

\section{Analysis Experiment Settings}
\label{sec:supp_analysis_settings}

All diagnostic experiments use the same answer extraction and accuracy computation as the main table: a prediction is correct only when the extracted option letter matches the final label. The implementation settings are:

\begin{itemize}
    \item \textbf{RQ1: Frame scaling.} We evaluate Qwen3-VL-8B and Lingshu-7B in the main figure, with MedGRPO included in auxiliary assets. The budgets are 4, 8, 16, 32, 64, 128, 256, and 512 uniformly sampled full-video frames.
    \item \textbf{RQ2: Evidence grounding.} We evaluate Qwen3-VL-4B, Qwen3-VL-8B, Lingshu-7B, and MedGRPO under random A/B/C/D, text-only, uniform vision+text, and GT evidence images+text. Uniform vision+text uses 64 frames, while the oracle setting uses the variable number of available evidence images for each item.
    \item \textbf{RQ3: Temporal order.} We evaluate Qwen3-VL-8B with a 256-frame budget under original order, reverse order, block shuffle with 8 blocks, and local shuffle with window size 16.
    \item \textbf{RQ3: Attention redundancy.} We analyze Qwen3-VL-8B and Lingshu-7B in the main figure, with MedGRPO included in auxiliary assets. The diagnostic uses 128 raw sampled frames resized for tractable attention extraction, then reports frame-level attention concentration, random-pair visual similarity, and patch-level redundancy around top-attention frame units.
    \item \textbf{RQ4: Sparse sampling.} We evaluate Qwen3-VL-8B at 32 and 64 frames under uniform sampling, differential frame-change sampling, and hybrid sampling with half uniform and half differential frames.
    \item \textbf{RQ4: Specialized long-video rows.} We evaluate AKS, ViLAMP, WFS-SB, VideoLLaMA3, LongVA, and VideoChat-Flash with the official budgets listed in Table~\ref{tab:open_protocol}, using each released method-specific selector or inference path.
\end{itemize}

\section{Reproducibility and Release Notes}
\label{sec:supp_reproducibility}

\paragraph{Evaluation code path.}
All local evaluations use sharded inference with resumable JSONL outputs. Each shard writes predictions incrementally and records model path, frame setting, batch size, retry count, and result file path in a manifest. OOM or worker failure can be resumed from the existing shard output without rerunning completed items. Closed-source runs are similarly checkpointed after each item to avoid losing completed API calls.

\paragraph{Compute.}
No model is trained for this paper. Local inference is run on institutional GPU servers with 8 A800 GPUs; large models use one or more GPUs depending on memory requirements. The dominant compute cost is inference over 1,253 long-video QA items and the frame-scaling/sampling diagnostic runs. Closed-source rows are reproduced through their public APIs, so hardware accounting is not controlled by the authors.

\paragraph{Responsible release.}
The release separates derived benchmark metadata from raw clinical videos. The package includes source identifiers, source URLs or reconstruction instructions, derived prompts/labels, task metadata, evidence metadata, and evaluation scripts. It does not claim new consent or a new license over upstream videos. Users must obtain raw videos from the original public source when redistribution is not permitted. The benchmark should be used for research evaluation and model diagnosis, not clinical decision making.

\clearpage

\end{document}